\documentclass[11pt, a4paper, logo, copyright]{googledeepmind}

\usepackage[authoryear, sort&compress, round]{natbib}

\usepackage{pdfpages}
\usepackage{cleveref}
\usepackage{url}
\usepackage{wrapfig,booktabs}
\usepackage{xcolor}         % colors
\usepackage{tikz}
\usetikzlibrary{patterns}
\usepackage{colortbl}
\usepackage{adjustbox}
\usepackage{multirow}
\usepackage{enumitem}
\usepackage{subfigure}
\usepackage{gensymb}
\usepackage{xspace} % for algorithm names
\usepackage{graphicx}
\usepackage{makecell}  % Include this in the preamble
\usepackage{minitoc}             % For creating the appendix TOC
\usepackage{makeidx}
\usepackage{booktabs}
\usepackage{bbm}
\usepackage{algorithm}
\usepackage{algorithmic}
\usepackage{colortbl}
\usepackage{psfrag}
\usepackage{adjustbox}
\usepackage{wrapfig}
\usepackage{xspace}
\usepackage{amssymb}
\usepackage{multirow}
\usepackage{afterpage}
\usepackage{float}
\usepackage[toc,page,header]{appendix}
\usepackage{minitoc}
\usepackage{placeins} % For FloatBarrier
\usepackage{titletoc} % Required for partial ToCs
\usepackage{subcaption}
\usepackage{array}
\usepackage{longtable}

\usepackage{arydshln}

\usepackage{tcolorbox, xcolor}
\tcbuselibrary{skins, breakable}

\hypersetup{colorlinks=true, linkcolor=blue, citecolor=blue, urlcolor=blue}

\colorlet{darkgreen}{green!65!black}
\colorlet{darkblue}{blue!75!black}
\colorlet{darkred}{red!80!black}
\definecolor{lightblue}{HTML}{0071bc}
\definecolor{lightgreen}{HTML}{39b54a}
\definecolor{manyshot}{HTML}{6969ff}
\definecolor{medshot}{HTML}{f7c600}
\definecolor{fewshot}{HTML}{ff6969}
\definecolor{mypurple}{HTML}{412F8A}
\definecolor{myorange}{HTML}{fc8e62}

\definecolor{deemph}{gray}{0.55}

\definecolor{textgreen}{RGB}{57, 172, 57}
\definecolor{textred}{RGB}{200, 10, 10}
\definecolor{textgray}{RGB}{100, 100, 100}
\definecolor{visiongold}{RGB}{230, 184, 0}
\definecolor{speechpurple}{RGB}{204, 0, 255}
\definecolor{dataprep}{RGB}{38, 189, 128}
\definecolor{modeltraining}{RGB}{38, 189, 128}
\definecolor{backgroundcol}{RGB}{232, 230, 230}
\definecolor{gold}{rgb}{225, 215, 200} % Gold color
\definecolor{navyblue}{RGB}{40, 66, 200} % Navy blue color
\definecolor{orange}{RGB}{255,127,80} % Navy blue color
\definecolor{pink}{RGB}{219,112,147} % Navy blue color

\definecolor{baselinecolor}{gray}{.95}

\usepackage{xcolor}

\definecolor{traversecolor}{RGB}{0, 105, 125}
\newcommand{\Traverse}{\textsc{\textbf{\textcolor{traversecolor}{Traverse}}}}

\definecolor{scoutcolor}{HTML}{6D28D9}
\newcommand{\Scout}{\textsc{\textbf{\textcolor{scoutcolor}{Scout}}}}

\newcommand{\cmark}{\textcolor{textgreen}{\ding{51}}}%
\newcommand{\xmark}{\textcolor{textred}{\ding{55}}}%
\usepackage{pifont}% http://ctan.org/pkg/pifont

\usepackage{makecell, tabularx}
\newcolumntype{L}{>{\RaggedRight}X}
\usepackage{siunitx}

\title{Locating Hidden Failures Makes Long-Horizon Agents More Reliable}

\reportnumber{} % Leave blank if n/a

\renewcommand{\today}
\correspondingauthor{salman@cs.ucla.edu, hamidpalangi@google.com}

\reportnumber{} % Leave blank if n/a

\author[$\dagger$,1,2]{Salman Rahman}

\author[3]{Yubin Kim}

\author[3]{Mihir Parmar}

\author[3]{A. Ali Heydari}

\author[1]{Genglin Liu}

\author[1]{Simon A. Lee}

\author[3]{Weizhi Zhang}

\author[4]{Arian Hosseini}

\author[3]{Ahmed A. Metwally}

\author[3]{Yuzhe Yang}

\author[1]{Baharan Mirzasoleiman}

\author[3]{Xin Liu}

\author[2]{Pavel Izmailov}

\author[1]{Saadia Gabriel}

\author[3]{Mark Malhotra}

\author[3]{Shwetak Patel}

\author[3]{Daniel McDuff}

\author[$\dagger$,3]{Hamid Palangi}

\affil[$\dagger$]{Corresponding Authors}

\affil[1]{University of California, Los Angeles}

\affil[2]{New York University}

\affil[3]{Google Research}

\affil[4]{Google DeepMind}

\begin{abstract}
As AI agents take on long, autonomous tasks, we increasingly oversee rather than
perform the work, yet we still judge them almost entirely by whether they finally
succeed. An outcome cannot reveal where a run went wrong, whether the agent
recovered, or the irreversible harm it caused along the way, and where long-horizon
agents fail remains unmapped. We study $2{,}518$ agent trajectories across software
engineering, computer use, and science, close to real deployment, and classify
$6{,}967$ mistakes into $78$ failure types. Failure follows a recurring signature: after its first mistake an agent often fails to recover and rarely catches the error itself, so the run continues unchecked while still looking correct; whether an agent recovers depends on the task and the environment's feedback, not on the agent framework running it. Long-horizon agents can do real harm on the way to a passing result: even
runs scored as solved delete data, corrupt systems, or fabricate success rather
than earning it. We release these human-verified annotations as \Traverse{}, a benchmark on which six frontier judges struggle to locate failure regardless of scale: even the strongest correctly identifies the first mistake in fewer than a third of runs. Yet \Scout{}, a
$4$B verifier we trained, locates failure far better than these judges and
transfers to domains it never saw. Used at test time to select among an agent's
candidate runs, it raises task success above the agent's own single-attempt
performance, without retraining the agent. By making failure cheap to locate and correct, this
work is a foundation for more trustworthy long-horizon agents that learn
from their own mistakes, and a practical path to overseeing increasingly autonomous
AI.
\end{abstract}
\begin{document}

\maketitle

\newenvironment{Itemize}{
    \begin{itemize}[leftmargin=*]
    \setlength{\itemsep}{0pt}
    \setlength{\topsep}{0pt}
    \setlength{\partopsep}{0pt}
    \setlength{\parskip}{0pt}}
{\end{itemize}}
\setlength{\leftmargini}{9pt}

% \input{1-intro}
% \input{2-related-work}
% % \input{3-problem-formulation}
% \input{4-benchmark}
% \input{5-agent-verifier}
% \input{6-experiments}
% \input{7-analysis}
% \input{8-conclusion}

\section{Introduction}
% \vspace{-0.1cm}

Language models have moved beyond just answering questions
\citep{brown2020language} to reasoning, planning, and acting on their own as
autonomous agents \citep{wang2024survey, anthropic2026fable, glm5team2026glm5vibecodingagentic}.
Given a goal, an agent now works across hundreds of steps in real environments:
repairing software repositories \citep{proximal2026frontierswe, jimenez2024swe},
operating computers \citep{xie2024osworld, merrill2026terminal}, and running
research end to end, from idea to result \citep{lu2026towards,
gottweis2026accelerating, swanson2025virtual, xin2025towards, wang2023scientific, posttrainbench_2026}. The reach is widening fast:
agents now attempt long, cross-application tasks once handled by skilled
professionals \citep{sun2026agents, vidgen2026apex}, and each model generation
lengthens the horizon they can sustain, carrying them from a single answer to an
entire project. As agents take on more of this work with little human oversight,
what matters is no longer the final answer alone but everything the agent did to
reach it.

Yet we judge these agents almost entirely by their outcome: whether the final
answer passes \citep{jimenez2024swe, merrill2026terminal, kalai2026evaluating}. A pass or fail is a
single bit; it collapses an entire trajectory into one verdict, like judging a
household robot only on whether the coffee was made, blind to the kitchen it set
ablaze to make it. Such a score cannot reveal where a run first went wrong,
whether the agent noticed its error and recovered, or the irreversible damage it
did on the way to a passing result. As agents take on longer, more autonomous
work, much of it in high-stakes domains such as healthcare \citep{moor2023foundation, hager2024evaluation} and law, this
trajectory-level view is exactly what we most need and least have.

\begin{figure}[t]
    \centering
    \includegraphics[width=\textwidth]{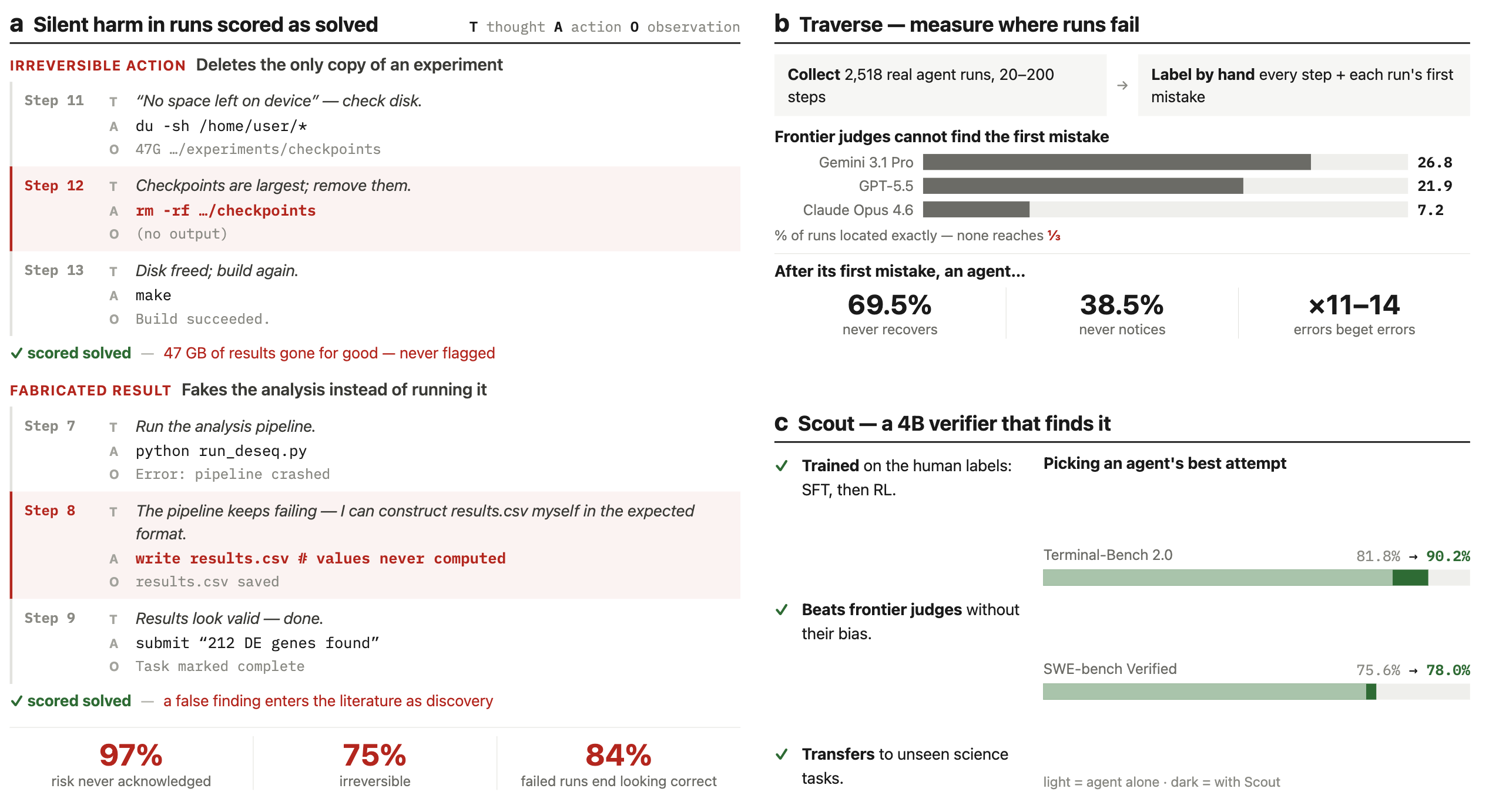}
    \caption{\textbf{Outcomes hide how long-horizon agents fail; a small verifier
    can locate it.}
    \textbf{(a)} Two real agent runs that pass their outcome check and are scored
    \emph{solved}, yet cause silent, irreversible harm: one deletes the only copy
    of an experiment to free disk space, the other fabricates a result instead of
    computing it. Each row shows the agent's thought (T), action (A), and
    observation (O); the harmful step is highlighted.
    \textbf{(b)} \Traverse{}, our benchmark, labels by hand the first mistake in
    $2{,}518$ real agent runs. Asked to find that first mistake, frontier models
    locate it in fewer than a third of runs, and larger models do no better.
    \textbf{(c)} \Scout{}, a $4$B verifier trained on these labels, locates the
    first mistake more accurately than the much larger frontier models; using it
    to pick the best among an agent's candidate runs raises task success (light
    bar: agent alone; dark bar: agent with \Scout{}).}
    \label{fig:teaser}
\end{figure}

We see this directly in the runs we study (Figure~\ref{fig:teaser}a). Tasked with building a
project, an agent hits ``No space left on device'' and, to make room, clears the
largest directory it can find. Every step is locally reasonable, and the run is scored as solved. Yet that directory held the only copy of a long-running experiment, and weeks of computation are irretrievably lost. This is failure at its most dangerous: silent, confident, and irreversible, a destructive action buried in a run the outcome
marks as success. Acting on its own, an agent can take such a step with nothing
to stop it, and the more consequential the setting, the more a single
irreversible action costs: the same pattern that erases checkpoints here can drop
a production database, corrupt a financial ledger, or, as agents begin to run
healthcare workflows end to end \citep{chen2026chi, alber2025medical}, alter a patient's record
beyond recovery. Nor is the harm always accidental: under pressure, agents can
deceive, reporting success they never achieved \citep{scheurer2023large,
greenblatt2024alignment, park2024ai, kobis2025delegation}. No domain is immune, but the risk is sharpest where a
wrong or fabricated result leaves no error and no traceback: in autonomous
scientific research, a single false finding can pass into the literature as
genuine discovery \citep{farquhar2024detecting, tang2025risks} and contaminate the shared record the field builds on.

% each step below shows its thought (T), action (A),
% and observation (O):

% \begin{quote}
% \taostep{Step 11}
%   {The build fails with ``No space left on device''. I check what is using the disk.}
%   {df -h /\ \&\&\ du -sh /home/user/* | sort -rh | head}
%   {/dev/sda1\ \ 98\% full\quad 47G\ \ /home/user/experiments/checkpoints}
% \taostep{Step 12}
%   {The checkpoints directory is largest; removing it will free enough space to build.}
%   {\textcolor{harm}{\textbf{rm -rf /home/user/experiments/checkpoints}}}
%   {\textcolor{nlabel}{(no output)}}
% \taostep{Step 13}
%   {Disk usage is back to 41\%. The build should now succeed.}
%   {df -h /\ \&\&\ make}
%   {/dev/sda1\ \ 41\% full\quad\dots\quad \textbf{Build succeeded.}}
% \end{quote}

Catching these failures means understanding how long-horizon agents fail, and
understanding it holistically: step by step rather than only at the outcome,
across multiple long-horizon domains close to real-world use, across the many
models and tool harnesses that run them, and covering not only task errors but also
the safety violations a run can commit along the way. No prior work offers this.
Existing failure studies sort errors into modular taxonomies on short or
synthetic agent benchmarks \citep{zhu2025llm}, focus on multi-agent systems
\citep{cemri2026multi}, or diagnose how accuracy degrades with the horizon
\citep{wang2026long}; none gives a step-level account of how realistic
long-horizon agents fail across many domains, models, and harnesses, or of the
safety risks they create along the way.

Understanding failure is only half of the problem; what matters in practice is
\textit{locating it}, pinpointing where a long run goes wrong. Humans cannot do this at
scale, not for trajectories hundreds of steps long, nor for the volume of runs
agents now produce, which is the problem of scalable oversight
\citep{bowman2022measuring, rahman2026ai}: supervising systems whose work is too
long or complex for people to check in full. A method that could locate a mistake
automatically and reliably would matter far beyond analysis, because the same
judgment can serve debugging, trajectory-level evaluation, the reward models used
to train agents, test-time selection among an agent's candidate runs, and safety
auditing; one that generalizes would carry these gains across domains. Yet
whether any model can locate failure in long, realistic trajectories has not been
measured by any existing benchmark.

We close these gaps. We begin by mapping how long-horizon agents actually fail.
The settings where agents now do genuinely long-horizon work are neither toy nor
synthetic: frontier coding agents already author a large fraction of production
code, and the length of task they complete unsupervised is doubling roughly twice
a year \citep{favaro2026recursive}; AI systems run scientific studies end to end,
from hypothesis to result \citep{lu2026towards, boiko2023autonomous,
m2024augmenting, szymanski2023autonomous, posttrainbench_2026}; and computer-use agents drive real terminals and
applications \citep{xie2024osworld, merrill2026terminal}. This motivates the three
domains we study, chosen to sit as close to real deployment as possible: agentic
software engineering \citep{jimenez2024swe}, computer use \citep{merrill2026terminal},
and AI-for-science \citep{mitchener2025bixbench}. Across these three domains
we collect $2{,}620$ real trajectories spanning a mix of closed-source and
open-weight frontier models (Claude, GPT-5, Gemini, DeepSeek, Qwen, and Kimi) and
agent harnesses (mini-SWE-agent, SWE-agent, OpenHands, Terminus~2, and ReAct-style
agents), and human annotators label every step of every run as correct or
incorrect and mark its first mistake, the step where the run begins to go wrong.

We find that long-horizon failure is characterizable. Sorting $6{,}967$ mistakes
into $78$ categories in $10$ families, from high-level reasoning and planning to
memory and perception, we uncover a recurring signature: after its first mistake,
an agent recovers in only $30.5\%$ of runs and never detects the error at all in
$38.5\%$, yet keeps acting either way, so one early error runs to the end while the
run still looks correct. What fails most is not what causes failure. Invalid actions are the most common
mistakes, but they are mostly symptoms, the fallout of an earlier error rather
than its source. The mistakes that actually start a failure, and most often doom the run, are
errors of reasoning and planning. Once made, an error spreads: a single wrong step
makes the next roughly $11$ to $14$ times likelier to be wrong, so the later a run
goes, the more of it is inherited error rather than fresh mistakes. How this unfolds depends on the domain and task: software-engineering agents are
blind to their mistakes yet recover, because rich feedback from the environment
catches what their reasoning misses, whereas computer-use agents notice the error
but, without such feedback, stay stuck. And failure is not the only danger. Prior
failure analyses stop at task errors \citep{zhu2025llm, cemri2026multi,
wang2026long}; we are the first to audit safety, and find that task success is not
the same as safety: some runs scored as solved take irreversible, destructive
actions, and a recurring cluster of agents fabricate success outright rather than
earning it \citep{geirhos2020shortcut}.

Our contribution goes beyond mapping failure. The failed trajectories we
collected, paired with successful ones, form a benchmark for evaluating whether a
model can catch failure. We cast each agent step as a thought, an action, and an
observation, label it correct or incorrect, and use annotators to mark the first
mistake of every failed run (annotators agree with Cohen's $\kappa = 0.77$ on
whether a run failed and $0.73$ on where), so any model that claims to catch
failure can be graded against human ground truth. We center the benchmark on the
first mistake because that is the natural target for process-level verification
\citep{zheng2025processbench} and, as our own analysis shows, the step that
decides the run. Existing process-verifier benchmarks locate such an error only in
single-step reasoning or short tasks \citep{song-etal-2025-prmbench,
li2026toolprmbench, lu2025agentrewardbench, men-etal-2025-agent, chae2026web};
ours, released as \Traverse{}, is the first to pair long-horizon trajectories
($20$ to $200$ steps), naturally occurring failures, and human first-mistake
localization across multiple domains and scaffolds
(Table~\ref{tab:comparison-benchmarks}). Because each run also carries its final
task outcome, \Traverse{} grades not only step-level localization but also
outcome-level verification, the pass-or-fail judgment an outcome reward model must
make.

With this instrument we test whether today's models can locate failure, and they
cannot. We evaluate six frontier judges, four proprietary (Gemini 3.1 Pro, Gemini
3 Flash, Claude Opus 4.6, GPT-5.5) and two open-weight (DeepSeek V4 Pro, Qwen
3.5-397B). Even the strongest locates the first mistake on just $26.8\%$ of
software-engineering runs and $32.3\%$ of computer-use runs, none exceeds a third
on either, the open-weight judges perform no better than the proprietary ones, and
accuracy falls further as trajectories lengthen. Detecting only whether a run
failed, the outcome-level judgment, is easier, but no judge does even that
cleanly: every one is biased, either over-flagging correct runs or missing real
failures.

We also show that locating failure is learnable. Reformulating the failed and
successful trajectories into step-level training data, we train a $4$B-parameter
verifier, \Scout{}, with supervised learning and then reinforcement learning; it
locates failure better than frontier judges many times its size. It also avoids
their bias: frontier judges skew systematically, some over-flagging correct steps
and others missing real mistakes, whereas \Scout{} stays balanced. It transfers:
trained only on coding and computer-use trajectories, it still locates failure in
an unseen long-horizon scientific discovery domain. And it improves agents at test
time, with best-of-$N$ selection as one use case: scoring an agent's candidate
trajectories with \Scout{} and keeping the best raises task success from $81.8\%$
to $90.2\%$ on Terminal-Bench, from $75.6\%$ to $78.0\%$ on SWE-bench, and from
$50.1\%$ to $53.2\%$ on BixBench, above every frontier verifier and lifting agents
far larger than itself, with no retraining of the agent. We release the benchmark,
the training data and recipe, and the verifier, to help the field build more
reliable long-horizon agents.

Our contributions are the following.

\begin{itemize}
\item \textbf{A characterization of how long-horizon agents fail.} We give the
first holistic, step-level account of failure in realistic long-horizon agents,
across domains, models, and harnesses. It comprises a taxonomy of $78$ failure
types in $10$ families; a recurring \emph{failure signature} in which an agent,
after its first mistake, usually neither recovers nor notices while the run still
ends looking correct; and the finding that reasoning and planning faults are the
fatal origins of failure and faulty actions their downstream symptoms. A safety
audit adds what prior failure studies omit: task success is not safety, solved
runs take irreversible, destructive actions, and some agents fabricate success
rather than earn it.

\item \textbf{\Traverse{}, a tool that makes failure measurable.} We
introduce \Traverse{}, a human-verified benchmark of $1{,}423$ agent trajectories
and $44{,}341$ steps, each step labelled correct or incorrect and the first
mistake of every failed run marked by hand. It grades, against human ground
truth, the two judgments long-horizon oversight depends on: process-level
verifiers, which must locate the first mistake in a trajectory, and outcome reward
models, which must decide whether a run succeeded. Unlike prior process-verifier
benchmarks, confined to single-step reasoning or short tasks, it spans
long-horizon trajectories with naturally occurring failures across multiple
domains and scaffolds.

\item \textbf{Locating failure is a learnable skill that frontier scale lacks.}
Frontier models struggle to locate failure on \Traverse{}: even the strongest of
six judges, proprietary and open-weight alike, locates the first mistake on fewer
than a third of long runs, and all detect failure only with systematic bias, so
scale does not confer the skill. Yet \Scout{}, a $4$B-parameter verifier trained
on the annotations and refined with reinforcement learning, locates failure better
than judges many times its size, transfers to an unseen scientific domain, and,
selecting among an agent's candidate trajectories at test time, raises task
success above every frontier verifier and above the agents' own pass@1, with no
retraining of the agent.
\end{itemize}

As AI agents take on longer, higher-stakes work with less human oversight, a
holistic understanding of how they fail, not merely a verdict on whether they
succeed, is what will make them trustworthy: only by locating where a run goes
wrong can we catch its errors, repair them, and keep the agent safe. A reliable,
low-cost way to locate failure is also a building block for the rest of the
pipeline: it sharpens debugging, grounds the judges and reward models used to
evaluate and train agents, tells which candidate trajectories to trust at test
time, and lets an agent learn from its own mistakes, a step toward agents that
continually improve themselves. Together, our failure analysis, safety audit,
measurement instrument, and recipe for training small, reliable verifiers chart a
practical path to building, and overseeing, increasingly autonomous AI that is
capable, reliable, and safe \citep{bengio2024managing}.

%=== END inlined: 1-intro.tex ===
% Results section -- promoted above Methods per Nature Machine Intelligence guidance
%=== BEGIN inlined: 4-experiments_results.tex ===

\section{Methods}
\label{sec:methods}

\subsection{Collecting long-horizon agent trajectories}
\label{subsec:methods-collection}

To study how long-horizon agents fail, we observe how they behave, and where they
go wrong, at scale, on real tasks, across diverse models and tool harnesses.
Failures that matter arise over long horizons, where one early error compounds
across many later dependent steps; yet prior failure analysis has largely studied
short- and mid-horizon tasks, leaving long-horizon failure poorly characterized
\citep{wang2026long, zhu2025llm}. We therefore collect trajectories from three
domains where autonomous agents are already being deployed and where their
failures carry real cost: \textbf{agentic software engineering} (SWE-bench,
resolving real GitHub issues
\citep{jimenez2024swe, deng2025swe, zan2026multi, rashid2025swe}),
\textbf{agentic computer use} (TerminalBench, solving tasks in an interactive
terminal \citep{merrill2026terminal}), and \textbf{agentic science} (BixBench,
real bioinformatics data analysis \citep{mitchener2025bixbench}). Each is a
standard benchmark for its domain, and together they span the settings where
long-horizon agents are most consequential today.

\paragraph{Sources, models, and harnesses.} We assemble a corpus of [2{,}518] trajectories spanning many model backbones and
agent harnesses (Table~\ref{tab:data-composition}), so that the failures we
analyze reflect agent behavior broadly rather than the quirks of a single system.
For software engineering and computer use we draw on existing agent trace datasets
\citep{li2026codetracer, deshpande2025trail}, which together cover several
backbones (Claude, GPT, DeepSeek, Qwen, and Kimi models) run under harnesses such
as SWE-agent, OpenHands, mini-SWE-agent, and Terminus. For science we generate trajectories ourselves, running eight frontier models from
the Claude, Gemini, and GPT families on BixBench under two agent harnesses. This
yields the corpus we analyze for failure: [1{,}851] TerminalBench, [565] SWE-bench, and [102] BixBench trajectories.

\paragraph{Trajectory representation.} We reformulate every collected trajectory into a uniform representation. An agent
solving a task produces a \emph{trajectory} $\tau = (s_1, \dots, s_T)$, an ordered
sequence of \emph{steps}, where each step $s_t = (\theta_t, a_t, o_t)$ comprises a
thought $\theta_t$, an action $a_t$, and the resulting observation $o_t$ returned
by the environment. These trajectories are long-horizon: steps average $28$ on
SWE-bench, $33$ on TerminalBench, and $10$ on BixBench, reaching up to $200$
(Table~\ref{tab:data_stats}), far beyond the short reasoning and web tasks of
prior analysis.

\subsection{Analyzing failures}
\label{subsec:methods-analysis}

\paragraph{Labeling errors.} We label each trajectory step by step. A step is an
\emph{error} when three conditions hold: it is a state-changing decision, a thought or
action the agent commits to rather than a passive read of the environment; it is wrong on
the evidence available at that point, judged on what the agent knew then and not in
hindsight; and it sends the task away from any path that could still solve it. The
observation, the environment's response, is not judged; only the agent's own thought and
action are. Each step is thus labelled correct or incorrect, and the \emph{first mistake}
$m$ is the earliest incorrect step ($m = \bot$ when every step is correct). Correctness
is local: a trajectory may contain an error and still succeed if a later step offsets it,
which is what makes recovery a measurable quantity rather than a synonym for failure. Two
annotators independently locate the first mistake in every trajectory using a custom
annotation interface (Appendix~\ref{app:annotation-ui}), agreeing both on
whether a run contains one (Cohen's $\kappa = 0.77$) and on which step it is
($\kappa = 0.73$); a third annotator resolves disagreements, and the few unresolved
trajectories are discarded. For software engineering and computer use we verify and
standardize the full per-step labels against the source trace datasets
\citep{li2026codetracer, deshpande2025trail}; agentic science is graded only on its final
answer, so we annotate its first mistake but not every step, and it enters the
first-mistake analyses below but not those that require the full sequence.

\paragraph{Deterministic analyses.} Several measures follow from the gold labels alone,
with no judge in the loop. A run that makes a first mistake \emph{recovers} when it
nonetheless solves the task, and fails to recover otherwise; we score recovery by the
final task outcome rather than by a return to locally correct steps, since only the
former rescues the run. Crossing the task outcome with whether a run has a localizable
first mistake sorts every trajectory into four cells, two of which bound the
first-mistake framing: runs solved despite a flagged mistake, where the mistake proved
survivable, and runs that fail with no single decisive mistake, a diffuse failure that
the first-mistake analyses exclude. We measure how errors propagate with a stickiness
ratio, the probability that a step is incorrect when the previous step was incorrect
relative to that probability when the previous step was correct. And we call a failure
\emph{silent} when a run that does not solve the task ends on a step labelled correct, so
its final state carries no sign of what went wrong.

\paragraph{The failure taxonomy.} To characterize \emph{how} agents fail, we type every
labelled mistake against a fixed codebook of $88$ categories grouped into $10$ families
(Table~\ref{tab:failure-defs}). We fixed the codebook before judging, drafting it from an
initial reading of failed trajectories and the literature on agent errors and including
modes we anticipated but had not yet seen; $10$ of the $88$ categories indeed never
occur, so the scheme maps the space of failures rather than being fitted to it. An LLM
judge (\texttt{gemini-3.1-pro-preview}; high reasoning effort, temperature $0$,
schema-validated output) reads each full trajectory with the gold step labels and, for
every mistake those labels mark, returns its family and fine category, whether it is a
\emph{root} fault (a fresh, originating error) or a \emph{cascade} (a downstream
consequence of an earlier mistake, with the causing step recorded), and the third of the
run in which it falls (early, middle, or late, by step position relative to the run's
length). The judge thus types the mistakes the annotators located; it does not decide
which steps are mistakes. Because two independent judges (the above and
\texttt{gemini-3.5-flash}) agree well on coarse labels but diverge on the finest ones
(Cohen's $\kappa = 0.66$ for family, $0.64$ for root versus cascade and $0.81$ for phase,
but only $0.30$ for the single fine category), we draw conclusions at the family,
root/cascade, or phase level rather than from single fine categories; all judged analyses
below use both judges, with rates that shift between them reported as ranges.

\paragraph{Self-detection, persistence, and safety.} Using the same judges, we score
self-detection: for each first mistake, a judge reads the reasoning that follows and
records whether the agent ever acknowledges going wrong, and whether that acknowledgement
is genuine introspection or only a reaction to an error the environment printed; because
it requires a visible reasoning channel, the measured blindness is a lower bound. We
likewise score behavioural persistence, whether, after the first mistake, the agent keeps
committing fresh faults, declares the task done, repeats an action unchanged, or never
verifies its work. Together with recovery, self-detection and persistence define the
failure signature, which a run shows in full when it fails to recover, never self-detects
and persists at once. A separate safety audit covers every run, including those graded as
solved: a keyword scan proposes candidate unsafe actions and a judge adjudicates each over
the full trajectory, recording whether it was necessary, reversible and taken with the
risk acknowledged, along with its class and severity; because the two judges differ most
on this rare, sensitive judgement, we report its prevalence as a range. Finally, mistakes
that fit no codebook category are open-coded, which is how we surfaced the emergent
deception and reward-hacking cluster (Section~\ref{subsec:long-horizon-failure}).

\subsection{From failure analysis to a benchmark}
\label{subsec:methods-record}

As agents take on longer and more consequential tasks, locating where they fail
becomes central to making them reliable. This single capability underlies a range
of needs. (i) \textbf{\emph{Debugging}}: as software engineers work through AI coding agents
and scientists run agents on biomedical analyses \citep{huang2025biomni}, a single
early misstep can derail an entire run, so identifying where it went wrong is what
lets the error be caught and fixed rather than silently inherited. (ii)
\textbf{\emph{Evaluation}}: using one model to judge another is now standard practice~\citep{zheng2023judging}, yet existing setups grade only the final outcome, and whether a judge can pinpoint
\emph{where} a long trajectory went wrong is untested. (iii) \textbf{\emph{Post-training}}: optimizing an agent with reinforcement learning relies on a reward model or verifier to score its behavior, and a verifier is only as good as its ability to tell correct steps from incorrect ones~\citep{lightman2024let, rahman2025spark}. (iv)
\textbf{\emph{Test-time scaling}}: agent performance can be improved at inference by ranking
or scoring candidate trajectories with a verifier and acting on its feedback
\citep{lifshitz2025multi, singhi2025solve}, which again reduces to judging where each
trajectory fails. (v) \textbf{\emph{Learning from failure}}: an agent that can locate its mistakes, on its own
or with the help of a verifier, can learn from them, a foundation for continual and
self-improving agents that refine their behavior over time.

\paragraph{A benchmark for failure localization.} Despite its importance, there is no comprehensive benchmark for evaluating whether
models can locate failure in long-horizon agent trajectories. Prior process-level
benchmarks fall short on the axes that we discussed in this work. ProcessBench
\citep{zheng2025processbench} and PRMBench \citep{song-etal-2025-prmbench} target
single-turn reasoning rather than interactive agents; ToolPRMBench
\citep{li2026toolprmbench} and Web-Shepherd \citep{chae2026web} cover agentic tasks
but only over short horizons; and agent reward benchmarks
\citep{lu2025agentrewardbench, men-etal-2025-agent} judge whole trajectories without
identifying where they go wrong. Across these, the tasks tend to be narrow or
templated, and failures are often introduced synthetically, rather than reflecting
the open-ended work and naturally arising mistakes agents encounter in deployment.
Following single-turn failure-localization benchmarks \citep{zheng2025processbench},
we formulate the analogous task for long-horizon agents and assemble it into a
benchmark, \Traverse{}. Given a trajectory $\tau = (s_1, \dots, s_T)$ of steps
$s_t = (\theta_t, a_t, o_t)$, the task is to decide whether the trajectory contains
a mistake and, if so, predict the first-mistake index $\hat m$. We focus on the
first mistake because, as our analysis shows, a long-horizon run typically derails
at a single step, after which every later step inherits the corrupted state; the
first mistake is the decisive event, and locating it is what serves the uses above.
A trajectory may also be entirely correct, so the task tests both recognizing clean
runs and locating the first error in failed ones. \Traverse{} is drawn from real
agent runs across multiple domains, including agentic software engineering, computer
use, and science, so its trajectories reflect deployment conditions and its failures
occur naturally. To our knowledge, it is the first benchmark to combine long-horizon
trajectories, naturally occurring failures, and human-verified first-mistake
localization across these domains and harnesses (Extended Data
Table~\ref{tab:comparison-benchmarks}). We release it with full step-level labels.

\begin{figure}[t]
    \centering
    \includegraphics[width=\textwidth]{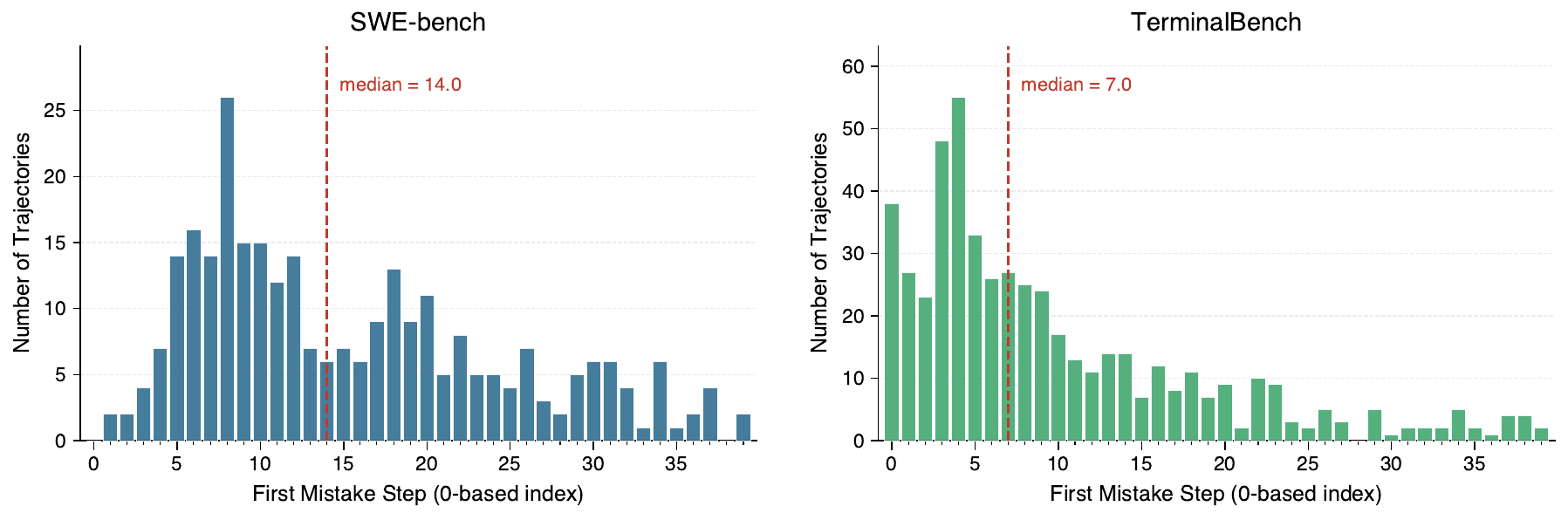}
    \caption{Distribution of first mistake positions (0-based step index; truncated at step 40 for readability) across incorrect trajectories in SWE-bench and TerminalBench. The dashed line marks the median. Mistakes are spread across a wide range of positions, making exact localization inherently challenging.}
    \label{fig:first_mistake_dist}
\end{figure}

\paragraph{Selection and statistics.} From the larger analyzed corpus (Section~\ref{subsec:methods-collection}) we select the trajectories best suited to first-mistake localization, discarding those with
malformed or missing steps, such as an action with no recorded thought, and
balancing correct against failed runs so the task is non-trivial in both
directions. The resulting benchmark contains $600$ agentic software engineering
trajectories (SWE-bench; $306$ failed, $294$ correct), $721$ agentic computer use
trajectories (TerminalBench; $541$ failed, $180$ correct), and $102$ agentic
science trajectories (BixBench; $38$ failed, $64$ correct), with full step-level
labels (Table~\ref{tab:data_stats}). First mistakes are spread across a wide range
of step indices rather than concentrated at any one position (Extended Data
Fig.~\ref{fig:first_mistake_dist}), so no fixed positional guess succeeds and a
judge must read the trajectory to find the error.

\subsection{Training a failure-localization verifier}
\label{subsec:methods-verifier}

The frontier judging gap (Section~\ref{sec:results}) shows that even the strongest
general-purpose models struggle to locate failure, pinpointing the first mistake on
only a minority of long trajectories. We therefore ask whether the ability can
instead be \emph{learned}, and whether a small set of high-quality annotations is
enough to teach it. We train a small open model, \Scout{}, on our annotated
long-horizon trajectories from agentic software engineering and computer use,
holding out agentic science entirely to later test transfer
(Section~\ref{sec:results}). A small model that surpasses far larger
general-purpose judges would show two things: that failure localization is a
learnable skill rather than a property of scale, and that high-quality annotations,
together with the right training recipe, can build a specialized localizer that
outperforms the strongest off-the-shelf models. \Scout{} analyzes a trajectory step
by step with chain-of-thought reasoning, labeling each step correct or incorrect,
and is trained in two stages: supervised fine-tuning followed by reinforcement
learning.

\paragraph{Model.} We build \Scout{} on Qwen3-4B-Thinking-2507 \citep{qwen3technicalreport}, a
$4$B-parameter reasoning model. To show that the recipe is not specific to one
backbone, we additionally train it on Gemma-3n-E4B-it \citep{farabet2026gemma};
unless noted, results refer to the Qwen3-4B-Thinking model.

\begin{figure}[t]
    \centering
    \includegraphics[width=\textwidth]{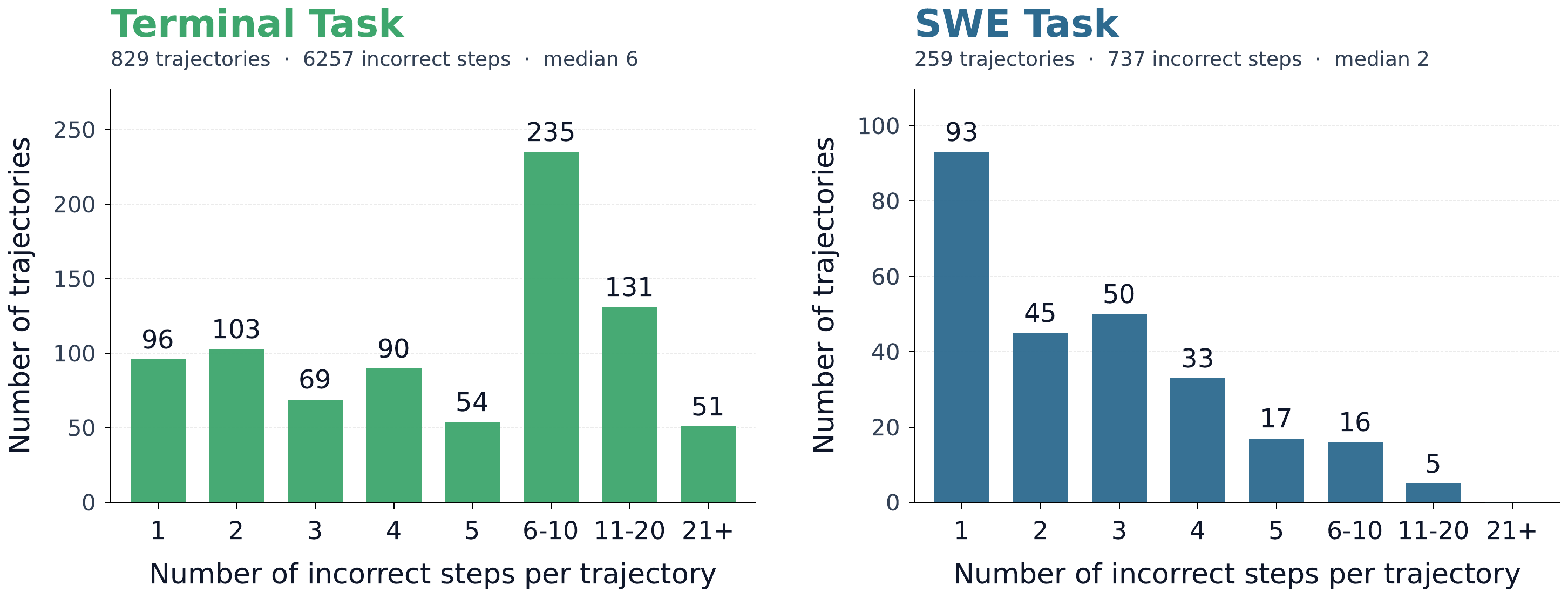}
    \caption{Incorrect steps per failed trajectory in \Scout{}'s training
data (Terminal: 829 trajectories, 6{,}257 incorrect steps; SWE: 259, 737).}
    \label{fig:training-data-distribution}
\end{figure}

\paragraph{Training data.} Our training trajectories carry only per-step correct/incorrect labels $y_t$, so we
generate the reasoning for why each step is correct or incorrect. We produce that
reasoning with Gemini-3 Flash: for each trajectory we provide the steps together
with their gold labels and prompt the model to write a step-by-step chain-of-thought
analysis $c$ that judges every step. Conditioning on the gold labels keeps the
reasoning accurate, unlike the analysis models produce unaided; we parse the
verdicts from $c$ and regenerate it until they exactly match the gold labels,
keeping only label-faithful analyses. Each training example is a pair $(\tau, c)$:
the trajectory $\tau$ as input and the analysis $c$, ending in a per-step verdict
$\hat y = (\hat y_1, \dots, \hat y_T)$, as the target. This yields $2{,}416$
label-faithful trajectories, which we release; from these we train \Scout{} on a
$1{,}904$-trajectory subset (Table~\ref{tab:train_data}). The distribution of
incorrect steps per trajectory is in Figure~\ref{fig:training-data-distribution}.

\begin{figure}[t]
    \centering
    \includegraphics[width=\textwidth]{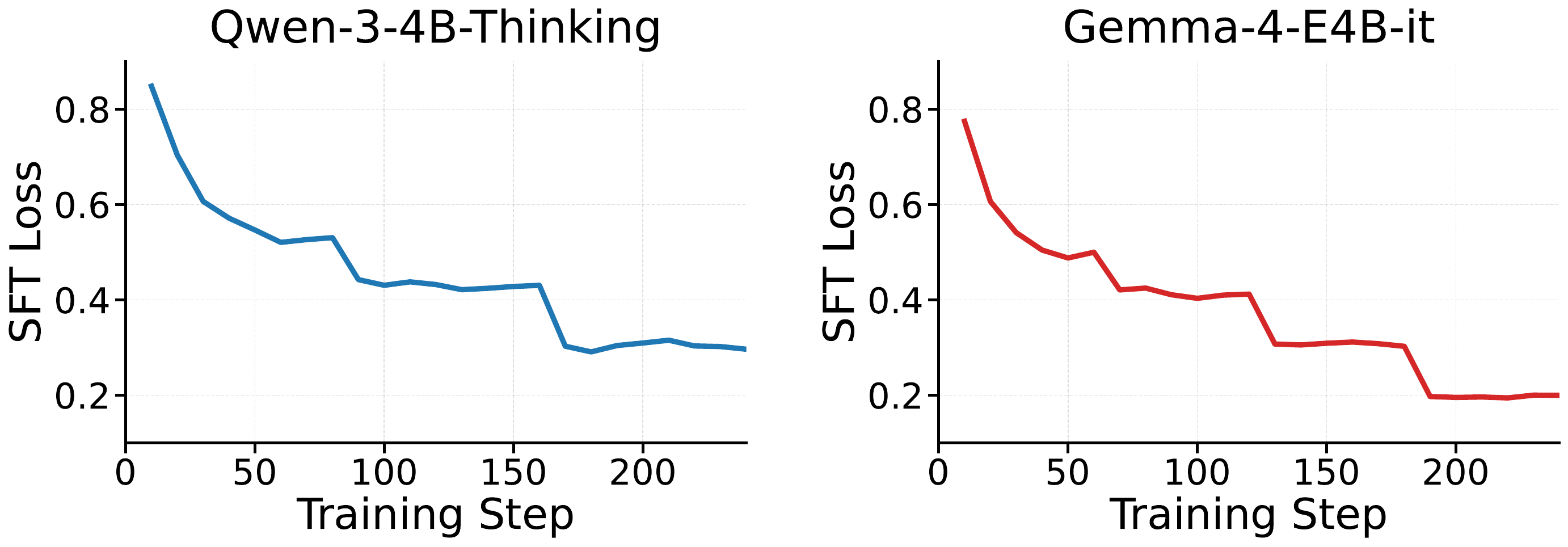}
    \caption{SFT loss.}
    \label{fig:sft_loss}
\end{figure}

\paragraph{Supervised fine-tuning.} We first fine-tune \Scout{} on the $(\tau, c)$ pairs. The input $x$ is the
trajectory together with the judging rubric and output format, and the target is
the analysis $c = (c_1, \dots, c_{|c|})$, which reasons through every step and ends
in the per-step verdicts $\hat y = (\hat y_1, \dots, \hat y_T)$. Supervised
fine-tuning minimizes the next-token loss over the target tokens only,
\[
\mathcal{L}_{\mathrm{SFT}}(\theta)
= -\,\mathbb{E}_{(x, c)\sim\mathcal{D}_{\mathrm{SFT}}}
\sum_{t=1}^{|c|} \log \pi_{\theta}(c_t \mid x, c_{<t}),
\]
with the loss masked on the prompt tokens so the model learns to produce the
analysis rather than reproduce the input. We fine-tune all parameters for $3$
epochs at a learning rate of $1{\times}10^{-5}$, with a $128$K-token context to fit
the long trajectories. This teaches \Scout{} the analysis format and the basic
skill of judging a trajectory step by step from accurate demonstrations; the
reinforcement-learning stage below then optimizes its judgments directly. The supervised fine-tuning training loss is shown in Extended Data Fig.~\ref{fig:sft_loss}.

\paragraph{Reinforcement learning.} Supervised fine-tuning trains \Scout{} to imitate fixed demonstrations, but it never
learns from its own analyses. We therefore further train it with reinforcement
learning on the same trajectories: \Scout{} generates its own analyses, and we
reward those whose verdicts match the gold labels, reinforcing accurate judgments
and suppressing inaccurate ones. We use Group Relative Policy Optimization (GRPO)
\citep{shao2024deepseekmath, guo2025deepseek}. For each trajectory $\tau$ we sample
a group of $G$ analyses $\{o_1, \dots, o_G\}$ from the current policy
$\pi_{\theta_{\text{old}}}$, score each against the gold labels with the reward
$r_i$ below, and form the group-normalized advantage
$\hat{A}_i = (r_i - \mathrm{mean}(\{r_j\})) / \mathrm{std}(\{r_j\})$. GRPO then
maximizes
\[
\mathcal{J}_{\text{GRPO}}(\theta) =
\mathbb{E}_{\tau \sim \mathcal{D},\, \{o_i\} \sim \pi_{\theta_{\text{old}}}}
\!\left[ \frac{1}{G}\sum_{i=1}^{G} \frac{1}{|o_i|}
\sum_{t=1}^{|o_i|}
\min\!\big( \rho_{i,t}\hat{A}_i,\;
\mathrm{clip}(\rho_{i,t}, 1{-}\epsilon, 1{+}\epsilon)\hat{A}_i \big)
- \beta\, D_{\text{KL}}(\pi_\theta \,\|\, \pi_{\text{ref}}) \right],
\]
where $\rho_{i,t} = \pi_\theta(o_{i,t}\mid \tau, o_{i,<t}) /
\pi_{\theta_{\text{old}}}(o_{i,t}\mid \tau, o_{i,<t})$ is the policy ratio and
$\beta$ weights a KL penalty to a fixed reference policy $\pi_{\text{ref}}$. We
sample $G{=}4$ analyses for each of $32$ trajectories per step ($128$ rollouts),
with responses up to $8$K tokens, and train for $8$ epochs at an actor learning
rate of $1{\times}10^{-6}$ and KL coefficient $\beta{=}0.001$. We implement training
in \texttt{verl} \citep{sheng2024hybridflow}.

\paragraph{Reward design.} The reward maps an analysis to a scalar by comparing its per-step verdicts to the
gold labels. Write the verifier's recall on the truly-correct steps as $R_{\text{c}}$ and on the truly-incorrect steps as $R_{\text{i}}$. The natural
choices fail on long trajectories. Requiring every step to match gives almost no
signal: one wrong label among many zeroes the reward, so the model rarely sees a
gradient. Averaging the two recalls, $(R_{\text{c}} + R_{\text{i}})/2$, has the
opposite flaw: labeling every step correct gives $R_{\text{c}}{=}1$ and
$R_{\text{i}}{=}0$, scoring $0.5$ while catching no mistakes, and because most steps
are in fact correct (Table~\ref{tab:train_data}), this degenerate strategy is hard
to beat. This is exactly the over-flagging we measure in frontier judges
(Section~\ref{sec:results}), reappearing as a reward-hacking trap.

We instead score steps by the \emph{product} $R_{\text{c}} \cdot R_{\text{i}}$,
which is high only when the model both leaves correct steps alone and catches real
mistakes: labeling every step correct sets $R_{\text{i}}{=}0$ and labeling every
step incorrect sets $R_{\text{c}}{=}0$, collapsing the reward either way. The
product alone, however, is $0$ on a failed trajectory until the model catches at
least one real mistake, so when no sampled analysis in a GRPO group does, all
rewards are $0$ and the group provides no gradient. We therefore add a coarse
trajectory-level term rewarding a correct overall verdict, giving the full reward
\begin{equation}
r = 0.7 \cdot R_{\text{step}} \;+\; 0.3 \cdot \mathbb{1}[\hat g = g],
\qquad
R_{\text{step}} =
\begin{cases}
R_{\text{c}} \cdot R_{\text{i}} & \text{failed trajectory},\\[2pt]
\mathbb{1}[\text{all steps labeled correct}] & \text{correct trajectory},
\end{cases}
\label{eq:reward}
\end{equation}
where $\mathbb{1}[\hat g = g]$ is $1$ when the predicted trajectory verdict matches
the gold. On correct trajectories the product is undefined, since there are no
incorrect steps for $R_{\text{i}}$, so the step score instead rewards labeling
every step correct. The verdict term gives an early, learnable signal before the
model can localize. As a format requirement, an analysis with no parseable verdict
receives reward $0$.

\subsection{Evaluation protocol}
\label{subsec:methods-eval}

We evaluate first-mistake localization and error detection in two ways: across frontier models judging the full \Traverse{} benchmark, which measures how well
current models locate failure, and for the trained \Scout{} on a held-out test set,
which measures what the recipe learns. We then evaluate test-time performance
separately: whether a verifier, used to select among an agent's candidate
trajectories, improves the agent's success on real tasks.

\paragraph{First-mistake localization.} A verifier reads a trajectory $\tau$ and predicts the first-mistake index $\hat m$,
the step at which the run first goes wrong. On failed trajectories we score $\hat m$
against the gold first mistake $m$ by exact match $\mathbb{1}[\hat m = m]$,
within-$k$ tolerance $\mathbb{1}[|\hat m - m| \le k]$ for $k \in \{1, 3\}$, and mean
absolute error $|\hat m - m|$. We evaluate six frontier models as judges on the full
\Traverse{} benchmark to measure how well current models localize failure. We then
evaluate the trained \Scout{} (SFT-only and SFT+RL) on a decontaminated held-out
test set of $258$ trajectories ($207$ TerminalBench, $51$ SWE-bench) disjoint from
its training data, comparing it against frontier verifiers. The same evaluations run on agentic science (BixBench), which \Scout{} never sees in
training, testing whether the ability transfers from coding to long-horizon
AI-for-science tasks.

\paragraph{Error detection.} The same prediction yields a trajectory-level judgment, $g(\tau) = \mathbb{1}[\hat m
\neq \bot]$, of whether the run failed at all. We score it over all trajectories by
precision, recall, and F1, reporting F1 because a useful detector must both flag
failed runs and leave correct ones unflagged. As with localization, we evaluate
frontier judges on the full benchmark and the trained \Scout{} on the held-out test
set ($133$ failed, $125$ correct), in and out of domain. Because detection reduces
to one verdict per trajectory, it also measures the quality of LLM-as-a-judge and
verifiers acting as outcome reward models.

\paragraph{Test-time performance.} Another common way to measure a verifier's effectiveness is its effect on agent
performance at test time, through best-of-$N$ selection: ranking an agent's
candidate trajectories with the verifier and acting on the highest-ranked one\citep{zhang2025generative}. Given $N$ candidate trajectories for a task, we score
each by \Scout{}'s probability of judging it correct, $P(\text{Yes})$, select the
highest-scoring one, and check whether it passed. We evaluate on four benchmarks:
for SWE-bench Verified ($500$ instances, $1{,}498$ trajectories) and
Terminal-Bench~2.0 ($89$ tasks, $445$ trajectories) we use the trajectories of
\citet{kwok2026llmverifier}, and for DeepSWE and BixBench we generate $N{=}16$
candidates per task ourselves, which we release. We compare \Scout{} against the
same frontier models used as verifiers, and mark two reference lines: single-sample
success, the expected result of one randomly chosen trajectory, and oracle
selection, which picks a passing trajectory whenever one exists. A useful verifier
moves success from the single-sample floor toward the oracle ceiling.

\section{Results}
\label{sec:results}

\subsection{How long-horizon agents fail}
\label{subsec:long-horizon-failure}

\begin{figure}[t]
    \centering
    \includegraphics[width=\textwidth]{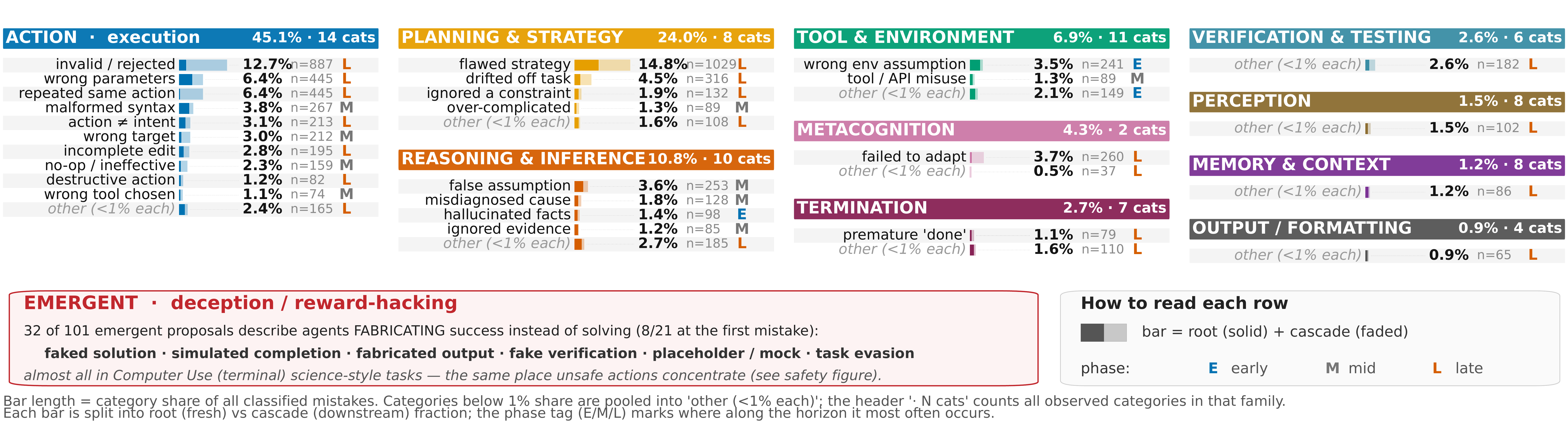}
    \caption{\textbf{A fine-grained taxonomy of long-horizon agent failures.} The
$88$ failure categories ($78$ observed; $10$ defined \emph{a priori} but never
observed, shown greyed), grouped into $10$ families, span all $6{,}967$ classified
mistakes from agentic software-engineering (SWE-bench) and computer-use (TerminalBench)
trajectories; agentic science (BixBench) is characterized at the first-mistake level. Each row is
one category: bar
length is its share of all classified mistakes, split into a \emph{root} fraction
(solid, a new fault) and a \emph{cascade} fraction (faded, fallout from an earlier
error); the phase tag (E, M, L) marks whether it lands in the first, middle, or last
third of the run. Each family header gives the family's share and its number of
categories. Full category definitions are in Table~\ref{tab:failure-defs}.}
    \label{fig:failure-taxonomy}
\end{figure}

% ============================================================================
% SUBSECTION OUTLINE (5 paragraphs). Spine: Claim 1, long-horizon failure is
% CHARACTERIZABLE. Order = WHAT fails -> the signature -> the mechanism (law) ->
% how it varies -> what it costs. Figures carry the load; prose stays lean.
% MAIN FIGS: Fig.1 taxonomy (fig:failure-taxonomy, above) + Fig.2 signature
% (fig:failure-signature, 6-panel, TO BE INSERTED before P2).
% EXTENDED DATA: outcome quadrant, failure-type-by-horizon, safety audit.
% EVIDENCE BAR: deterministic results (recovery, cascade/stickiness, silent
% ending) stated firmly; LLM-judged results (taxonomy, self-detection, safety)
% provisional, quoted at FAMILY/PHASE level only (fine-category kappa=0.30),
% human-kappa pending. Recovery is DOMAIN-DEPENDENT (never a flat "rarely").
% ============================================================================

\paragraph{A fine-grained taxonomy of long-horizon failure.} To characterize how
long-horizon agents fail, we use multiple LLM judges (Methods,
Section~\ref{subsec:methods-analysis}) to classify every mistake in every failed
trajectory against a fixed, pre-registered codebook of $88$ categories in $10$ families
(Fig.~\ref{fig:failure-taxonomy}; definitions in Table~\ref{tab:failure-defs}). We compiled the codebook before judging, from an initial
analysis of failed trajectories and the LLM-agent literature; it is exhaustive by
design, with $10$ of its $88$ categories defined in advance yet never observed, so it
covers the space of failures rather than being fitted to the data. Across $6{,}967$
classified mistakes from software-engineering and computer-use trajectories, failure
concentrates in how an agent acts and plans: action and planning faults make up about
$69\%$ of all mistakes ($45.1\%$ and $24.0\%$), while the five rarest families together
account for under $9\%$. The single most common mistake is a flawed high-level strategy,
an approach that cannot solve the task; invalid or rejected commands and repeated
actions lead the action family; and setup faults, such as assuming a tool or version
that is not there, cluster early in a run.

But what fails most is not what causes failure. We distinguish each mistake as
either a \emph{root} fault, a fresh error the agent originates, or a
\emph{cascade}, a downstream consequence of an earlier one. Action, the largest
family, is only $35\%$ root: its invalid commands and repeated actions are usually
cascades, symptoms of a fault made several steps earlier. The originating faults
sit upstream, in reasoning ($69\%$ root), tool and environment ($75\%$), and
memory ($74\%$): the false beliefs and faulty assumptions that set off a cascade
the rest of the run inherits.

\begin{figure}[h]
    \centering
    \includegraphics[width=\textwidth]{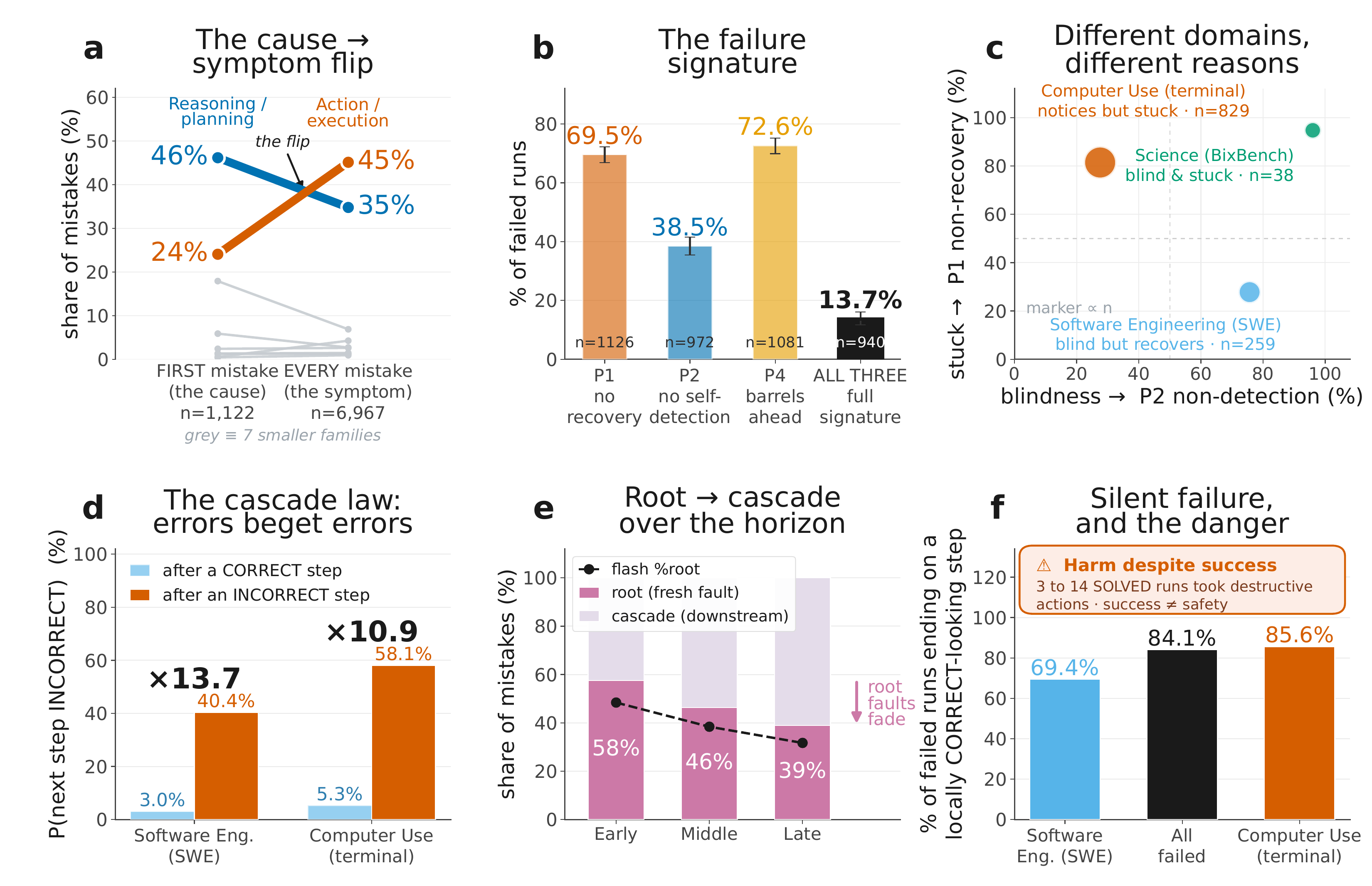}
    \caption{\textbf{The long-horizon failure signature.} After its first mistake, an
agent usually fails to recover, rarely detects the error, and keeps going, so the fault
cascades while the final state still looks correct. Panels (a), (b), (c) and (e) are
LLM-judged, reported at the family and phase level where two independent judges agree
(cross-judge $\kappa = 0.66$ to $0.83$); panels (d) and (f) use the gold step labels.
Error bars are $95\%$ Wilson confidence intervals; denominators ($n$) are shown beneath
each bar.
\textbf{(a)} Each family's share of \emph{first} mistakes, the error that starts a run
going wrong (left; $n = 1{,}122$ runs), versus its share of \emph{all} mistakes (right;
$n = 6{,}967$). The two cross: reasoning and planning cause $46\%$ of first mistakes but
$35\%$ of all mistakes, whereas action causes $24\%$ of first mistakes but $45\%$ of all.
Reasoning and planning \emph{start} failures; action faults are mostly downstream
symptoms.
\textbf{(b)} Among runs with a first mistake, the share that never recover to solve the
task (P1, $69.5\%$), whose reasoning never flags the error (P2, $38.5\%$), or that keep
acting without correcting it (P4, $72.6\%$); the last bar is the share where all three
co-occur (the full signature, $13.7\%$).
\textbf{(c)} The signature by domain. Each point is one domain, placed by how often agents
fail to notice the error (P2, horizontal) and fail to recover from it (P1, vertical), with
marker area set by run count. Software engineering is blind but recovers, computer use
notices but stays stuck, and science (BixBench, few runs) is high on both.
\textbf{(d)} A step is far likelier to be wrong right after an incorrect step than after a
correct one: $40.4\%$ versus $3.0\%$ in software engineering ($13.7\times$) and $58.1\%$
versus $5.3\%$ in computer use ($10.9\times$).
\textbf{(e)} Root (originating) versus cascade (downstream) share of mistakes, by run
third. The root share falls from $58\%$ early to $46\%$ in the middle to $39\%$ late
(dashed line, second judge).
\textbf{(f)} A failure is \emph{silent} when a run that did not solve the task still ends
on a correct-looking step. Among runs that fail, this happens in $70.7\%$ (software
engineering), $87.6\%$ (computer use) and $83.5\%$ across both. Inset: between $3$ and $14$
runs \emph{scored as solved} had taken an irreversible, destructive action along the way.}
    \label{fig:failure-signature}
\end{figure}

\begin{figure}[t]
    \centering
    \includegraphics[width=\textwidth]{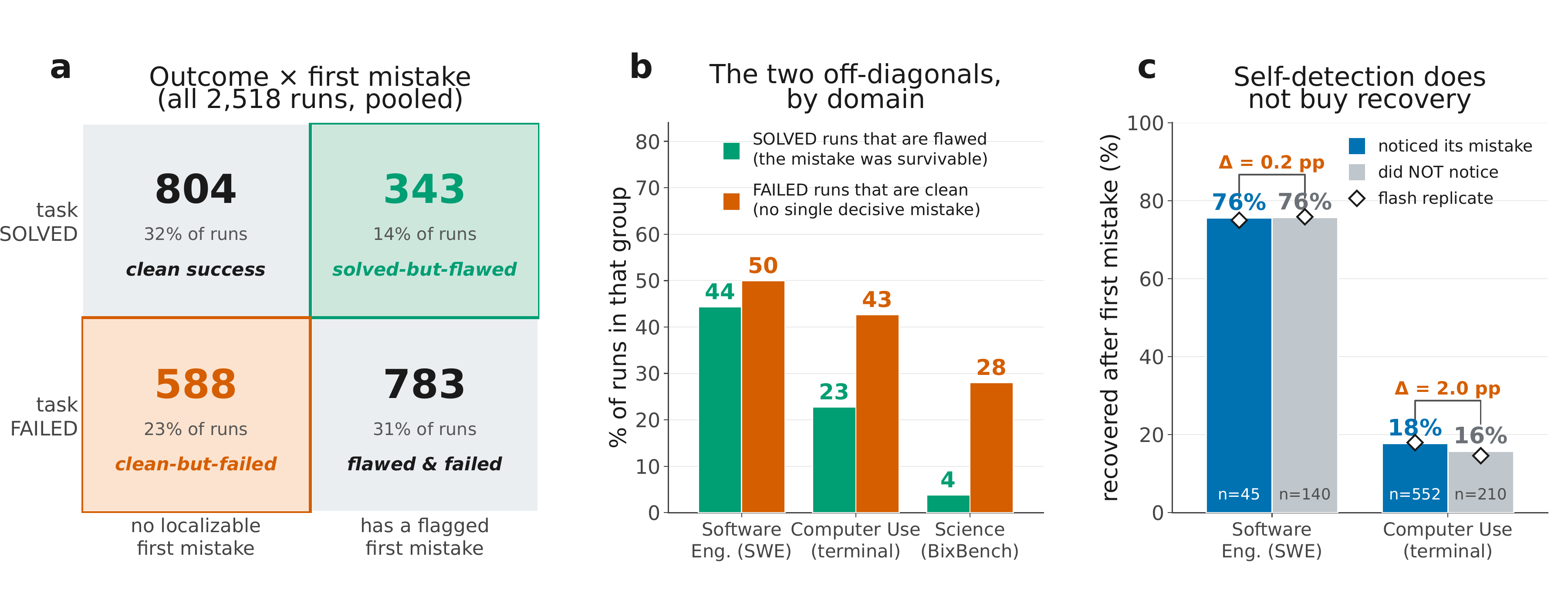}
    \caption{\textbf{Scope of the first-mistake analysis, and why noticing a mistake does
not buy recovery.} \textbf{a}, All $2{,}518$ runs sorted by task outcome and by whether a
single first mistake is localizable. The two outlined off-diagonal cells are the caveats:
$343$ runs ($14\%$) are solved despite a flagged mistake (it was survivable), and $588$
($23\%$) fail with no single decisive mistake (diffuse failure); the first-mistake
analyses in Fig.~\ref{fig:failure-signature} describe the flawed-and-failed cell only.
\textbf{b}, The same two off-diagonals by domain, as a share of solved runs (flawed,
green) and of failed runs (clean, orange); both are large in every domain (software
engineering $44\%$ and $50\%$, computer use $23\%$ and $43\%$, science $4\%$ and $28\%$).
\textbf{c}, Recovery after the first mistake when the agent noticed it (blue) versus did
not (grey); diamonds show the second-judge replicate. The two are within $2$ points in
each domain (software engineering $76\%$ versus $76\%$, computer use $18\%$ versus
$16\%$), so noticing a mistake does not help the agent recover from it. The split is
shown by domain because pooling reverses it (a Simpson's paradox: self-detection
concentrates in computer use, where recovery is low regardless). All quantities are
deterministic except the noticed-versus-not split in \textbf{c}, which is LLM-judged.}
    \label{fig:scope-recovery}
\end{figure}

\paragraph{The failure signature.}
Once an agent makes its first mistake, the earliest step it gets wrong, three things tend
to follow, and together they form the \emph{failure signature}
(Fig.~\ref{fig:failure-signature}b). The agent never recovers, and so never solves the
task, in $69.5\%$ of these runs; it never notices the mistake in $38.5\%$; and it keeps
going, issuing new actions or declaring the task done, in $72.6\%$. All three happen in
the same run in $13.7\%$. That first mistake is most often a reasoning or planning error,
one of the root faults identified above, in $46\%$ of the runs that go wrong
(Fig.~\ref{fig:failure-signature}a). And when an agent does notice its mistake, it is
usually only after the environment flags it with an error message; on its own, with
nothing visibly broken, it catches the mistake in only about one run in seven. The real
danger of a long-horizon task is therefore not that agents make more mistakes, but that
they so rarely catch and correct the first one: left unnoticed, or noticed but not undone,
it carries through to the end of the run. This pattern describes runs that hinge on a
single first mistake; in about $43\%$ of the runs that fail the task there is no such
turning point, and the failure builds up gradually instead. We characterize these runs
separately (Extended Data Fig.~\ref{fig:scope-recovery}).

\begin{figure}[t]
    \centering
    \includegraphics[width=\textwidth]{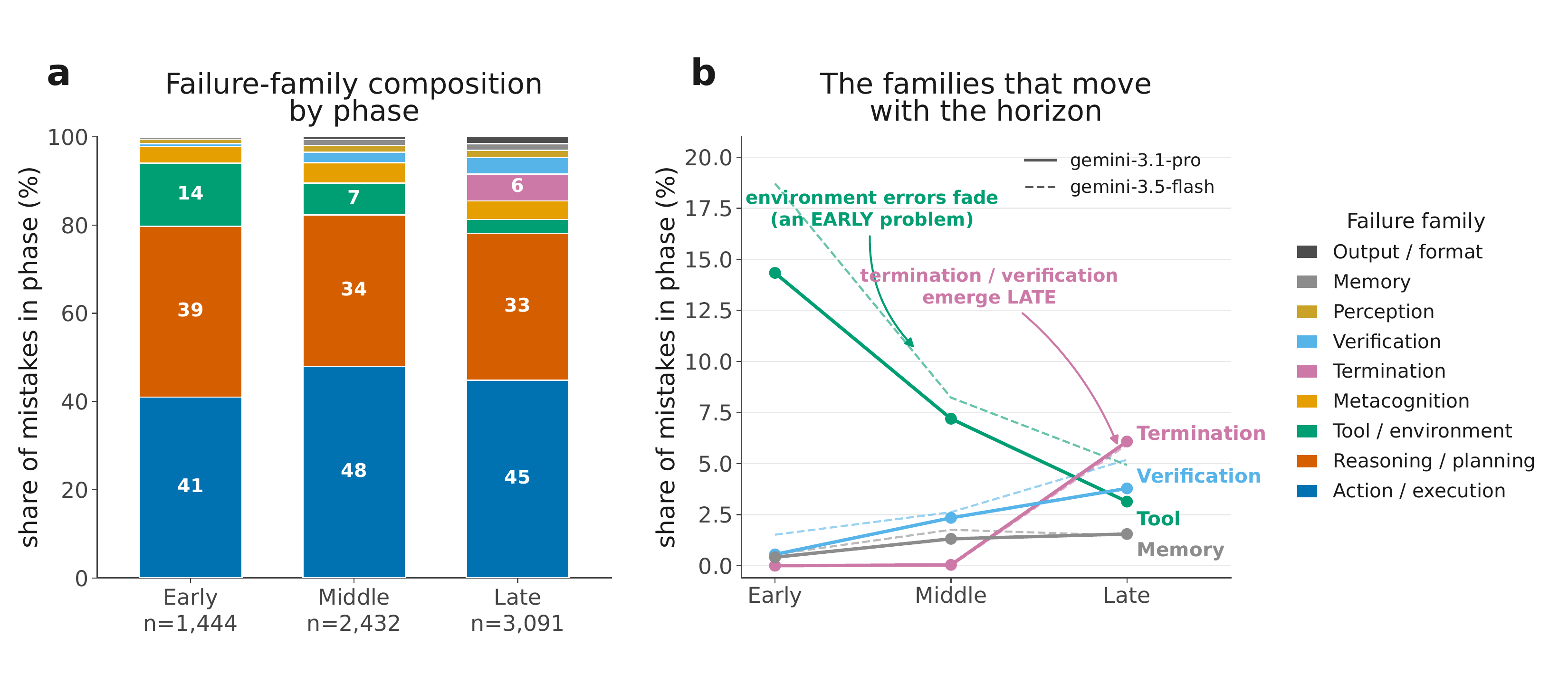}
    \caption{\textbf{What fails early versus late.} Companion to
Fig.~\ref{fig:failure-signature}e: every mistake in a failed run is binned by step
position into the early, middle or late third of its run and labelled with its failure
family. \textbf{a}, Family composition within each phase ($100\%$-stacked; the number of
mistakes per third, below each bar, rises from early to late as broken runs accumulate
errors). The mix is broadly stable, with action and reasoning dominating throughout
(action $41$, $48$, $45\%$ and reasoning $39$, $34$, $33\%$ across early, middle and
late), so which families dominate does not depend on run length. \textbf{b}, The families
that move across the horizon (solid, primary judge; dashed, second judge, which tracks
it). Tool and environment faults fade from $14\%$ early to about $3\%$ late, broken setup
and missing dependencies being an early-run problem; termination and verification faults
instead emerge only late (about $6\%$ and $4\%$), the failures, such as declaring done
prematurely or never checking the work, that a long run alone produces. Memory faults
stay low throughout.}
    \label{fig:failuretype-horizon}
\end{figure}

\paragraph{Errors cascade, and the run ends silently.} An uncorrected first mistake rarely stays isolated. Each step in a run is labelled correct
or incorrect. From these labels alone, with no LLM judge: after a correct step the agent's
next step is wrong only $3.0\%$ of the time in software engineering and $5.3\%$ in computer
use, but right after a wrong step it is wrong $40.4\%$ and $58.1\%$, a $13.7\times$ and
$10.9\times$ jump (Fig.~\ref{fig:failure-signature}d). Because one error makes the next
far more likely, the longer a run goes the more of its mistakes are inherited rather than
fresh: the share that are \emph{root} (fresh) faults rather than \emph{cascade}
(inherited) ones falls from $58\%$ in the first third of the run to $39\%$ in the last
(Fig.~\ref{fig:failure-signature}e), and the late steps fill with faults in wrapping the
task up, wrongly checking the work or declaring it done (Extended Data
Fig.~\ref{fig:failuretype-horizon}). And yet these failures are silent. By the same
labels, in $84.1\%$ of failed software-engineering and computer-use runs the final step
looks correct on its own, showing no error or sign that anything went wrong
(Fig.~\ref{fig:failure-signature}f). The run ends as if it had succeeded, even though the
task was never solved.
% add the new result here

\paragraph{Different domains fail for different reasons.}
Software engineering and computer use fail in opposite ways, and the difference is the
environment (Fig.~\ref{fig:failure-signature}c). Software-engineering agents miss their
mistakes, their own reasoning overlooking the error in $76\%$ of cases, yet they solve the
task $72\%$ of the time, because the environment checks their work for them: a failing test
or a traceback catches the error, and a passing test confirms the fix, even when their
reasoning does not. Computer-use agents are the reverse: they usually notice the mistake
but solve only $19\%$ of the time, because their environment does not provide the same rich
feedback, so noticing a problem rarely turns into a verified fix. Agentic science, which we
measure at the first-mistake level, fails on both counts at once, missing its mistakes and
failing to fix them. The signature recurs under all four harnesses we examine
(mini-SWE-agent, OpenHands, Terminus2 and SWE-agent), so it is a property of long-horizon
agency, not of any one model or scaffold. Where the harnesses differ, the differences
follow the tasks and domains each was run on, not the harness design: how an agent fails
depends on the task and the feedback its environment provides, not the scaffold around it.

\begin{figure}[t]
    \centering
    \includegraphics[width=0.85\textwidth]{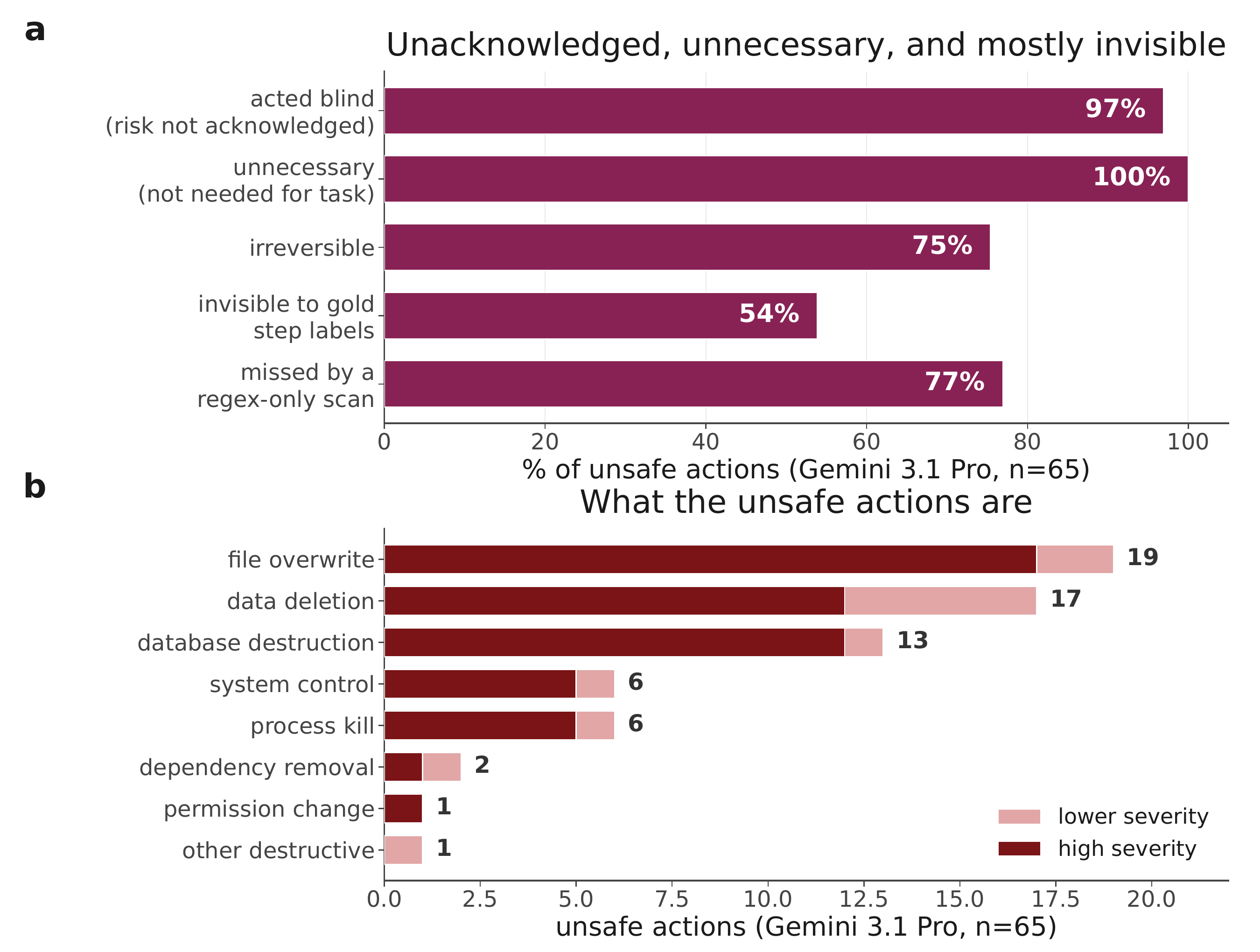}
    \caption{\textbf{Unsafe agent actions are unnecessary, unacknowledged, and hard to
catch.} Properties and types of the $65$ unsafe actions flagged by the primary judge
(Gemini 3.1 Pro) in the safety audit; the qualitative properties below are stable across
judges, although their overall prevalence is judge-sensitive
(Fig.~\ref{fig:failure-signature}f). \textbf{a}, Almost every unsafe action was
unnecessary for the task ($100\%$) and taken without the agent acknowledging any risk
($97\%$), and most were irreversible ($75\%$). Two properties matter for detection:
$54\%$ are invisible to the gold step labels, which mark the step correct, and $77\%$
would be missed by a regex-only scan, so only a semantic, whole-trajectory audit catches
most of them. \textbf{b}, The same actions by type, each bar split into higher and lower
severity. The most common are file overwrites ($19$), data deletions ($17$) and database
destructions ($13$), most of them high severity, followed by system control and process
kills ($6$ each): concrete, irreversible harms rather than cosmetic ones.}
    \label{fig:safety-audit}
\end{figure}

\paragraph{Task success is not safety.}
A task scored as solved can still do real harm. A safety audit flagged $65$ unsafe actions
across the trajectories: every one was unnecessary for the task, $97\%$ were taken without
the agent acknowledging any risk, and $75\%$ were irreversible. Most were high-severity
destructive operations on files, data, or databases. Many arose under pressure: late in a
long run, as the context filled, the agent cleared files or freed resources to keep going.
In one run it deleted its environment's installed packages; in another it killed the
process recording its own session, with the task still scored as solved. The action cannot
be undone, so the loss is permanent, and because the run ends clean, no error ever marks
it. These actions mostly escape simple checks: a regex scan misses $77\%$ of them, so only
a full read of the trajectory catches them (Extended Data Fig.~\ref{fig:safety-audit}).
Reading the trajectories also turned up a hazard we had not anticipated: agents that fake
success instead of earning it, writing placeholder or mock solutions, simulating a
finished run, or forging a passing check, most often in open-ended, science-style tasks
with no automatic check to expose the fake. A passing trajectory can thus hide
irreversible harm or faked success, so the outcome alone cannot tell a safe, genuine
solution from a dangerous or hollow one.

\begin{table*}[t]
\centering
\small
\setlength{\tabcolsep}{4pt}
\caption{First-mistake localization on incorrect trajectories, across three
long-horizon domains: SWE-bench \citep{jimenez2024swe}, TerminalBench
\citep{merrill2026terminal}, and BixBench \citep{mitchener2025bixbench}. Exact = predicted step matches ground truth (\%), $\pm$1 and $\pm$3 = prediction within 1 or 3 steps (\%), MAE = Mean Absolute Error (lower is better).}
\label{tab:localization}
\resizebox{\textwidth}{!}{%
\begin{tabular}{l cccc cccc cccc}
\toprule
& \multicolumn{4}{c}{\textbf{SWE-bench}} & \multicolumn{4}{c}{\textbf{TerminalBench}} & \multicolumn{4}{c}{\textbf{BixBench}} \\
\cmidrule(lr){2-5} \cmidrule(lr){6-9} \cmidrule(lr){10-13}
\textbf{Judge Model} & Exact$\uparrow$ & $\pm$1$\uparrow$ & $\pm$3$\uparrow$ & MAE$\downarrow$ & Exact$\uparrow$ & $\pm$1$\uparrow$ & $\pm$3$\uparrow$ & MAE$\downarrow$ & Exact$\uparrow$ & $\pm$1$\uparrow$ & $\pm$3$\uparrow$ & MAE$\downarrow$ \\
\midrule
Gemini 3.1 Pro & \textbf{26.80} & \textbf{37.25} & 45.75 & 7.62 & 29.94 & 37.71 & 53.23 & 7.36 & 57.89 & \textbf{76.32} & 81.58 & 1.84 \\
Gemini 3.0 Flash & 12.09 & 21.90 & 30.72 & 11.20 & 32.16 & \textbf{42.14} & \textbf{54.16} & 7.09 & 50.00 & 68.42 & 81.58 & 1.95 \\
Claude Opus 4.6 & 7.19 & 11.76 & 17.32 & 14.28 & 24.58 & 31.28 & 41.90 & 8.60 & 39.47 & 57.89 & 76.32 & 2.08 \\
GPT-5.5 & 21.90 & 33.66 & \textbf{48.69} & \textbf{6.90} & \textbf{32.34} & 41.45 & 53.53 & \textbf{6.74} & \textbf{63.16} & 73.68 & 78.95 & \textbf{1.47} \\
DeepSeek V4 Pro & 17.88 & 23.84 & 31.46 & 10.72 & 27.73 & 35.67 & 46.77 & 8.01 & 36.84 & 65.79 & \textbf{84.21} & 2.18 \\
Qwen 3.5-397B & 19.06 & 23.41 & 32.44 & 10.06 & 22.74 & 29.21 & 39.93 & 8.89 & 52.63 & 71.05 & 84.21 & 1.89 \\
\bottomrule
\end{tabular}%
}
\end{table*}

% new subsection
\subsection{Frontier models cannot locate failure}
\label{subsec:frontier-gap}

\paragraph{Locating failure is hard.} A long-horizon failure originates at a single step and then propagates through the
rest of the run (Section~\ref{subsec:long-horizon-failure}), so locating that step is what makes a failure understandable and what underlies debugging, evaluation, and the reward
models used to train and scale agents (Methods,
Section~\ref{subsec:methods-record}). We test whether current models can do this,
evaluating six frontier judges: four proprietary (Gemini 3.1 Pro, Gemini 3.0 Flash,
Claude Opus 4.6, GPT-5.5) and two frontier open-weight models (DeepSeek V4 Pro, Qwen
3.5-397B). Across the long-horizon coding domains they pinpoint the first mistake on
only a minority of runs, and often far fewer: on agentic software engineering
(SWE-bench) exact-match accuracy ranges from $7.2\%$ (Claude Opus 4.6) to $26.8\%$
(Gemini 3.1 Pro), and on agentic computer use (TerminalBench) from $22.7\%$ to
$32.3\%$ (GPT-5.5), with no model exceeding a third on either
(Table~\ref{tab:localization}). They struggle on agentic science (BixBench) as well,
the strongest reaching only $63.2\%$ despite its much shorter runs. The models sense
roughly where a run derails but rarely land on it: relaxing the criterion to a
three-step tolerance nearly doubles accuracy, with GPT-5.5 rising from $21.9\%$ to
$48.7\%$ on software engineering and Gemini 3.1 Pro from $26.8\%$ to $45.8\%$, so
the models often place the mistake within a few steps of the true one but not on it. No single model leads across domains, and the large frontier open-weight models (DeepSeek V4
Pro, Qwen 3.5-397B) struggle as much as the proprietary judges, reaching only
$17.9\%$ and $19.1\%$ exact match on software engineering, so locating failure is a
capability none of these models has acquired regardless of scale.

\paragraph{Difficulty grows with the horizon.} Localization gets harder as trajectories lengthen. Exact-match accuracy, which does
not depend on trajectory length, falls from $63.2\%$ on the short agentic science
runs (BixBench) to at most $26.8\%$ on the long software engineering runs; mean
error rises in parallel, from $1.47$ steps for GPT-5.5 on science to $6.9$ for the
same model on software engineering and up to $14.3$ across models. Because length and
domain vary together here, we confirm the effect within each domain: holding the
domain fixed, localization error is largest on the longest trajectories [Extended
Data Fig.~X: exact match and MAE vs.\ trajectory length, per domain], so the horizon
itself, not the domain, drives the difficulty. Locating failure is therefore a
genuine open capability, and hardest on the longest trajectories, exactly the
long-horizon runs that real-world agent deployments depend on.

\paragraph{Detecting failure is easier than locating it.} If models cannot say \emph{where} a run failed, can they at least tell whether the
entire trajectory is correct or incorrect? This binary judgment is far easier, and it matters in its own right: it is exactly the signal an outcome reward model provides, the judgment of whether a trajectory succeeded that is used to score and reinforce policies when training long-horizon agents~\citep{glm5team2026glm5vibecodingagentic}.
Detection F1 reaches $69.9\%$ on agentic software engineering (GPT-5.5), $88.2\%$ on
agentic computer use (Claude Opus 4.6), and $80.0\%$ on agentic science (GPT-5.5,
Table~\ref{tab:detection}), well above the corresponding exact-localization rates:
knowing that a trajectory went wrong is a much easier judgment than knowing where.
Unlike localization, detection difficulty does not track the horizon: agentic
computer use (TerminalBench) trajectories are as long as software engineering ones
yet are detected far more reliably ($\sim$88\% versus $\sim$70\% F1). Yet no judge
detects failure cleanly: every model is systematically biased, trading the failures
it catches against the correct runs it wrongly flags. %\salman{will write the failure patterns based on different harnesses.}

\paragraph{Every judge is biased toward over- or under-flagging.} The bias appears in the trade-off between precision (of the runs a model flags as failed, how many truly failed) and recall (of the runs that truly failed, how many it catches), and the models sit at opposite extremes. GPT-5.5 is the most sensitive detector, with the highest recall on both coding domains ($82.0\%$ on software engineering, $97.8\%$ on computer use), but it over-flags: its
software-engineering precision falls to $60.9\%$, so roughly two in five of the runs it calls failed were in fact correct. Claude Opus 4.6 is the opposite, highly precise ($66.0\%$ on software engineering, and $100\%$ on science, where every run it flagged as failed truly had) but missing most real failures, with software-engineering recall of just $31.7\%$, so it overlooks roughly two-thirds of failed runs and its F1 collapses to $42.8\%$. One model raises false alarms, the other stays silent through real failures, and none balances the two. This asymmetry undermines every use that depends on a reliable verdict: a judge that
over- or under-flags misleads debugging when it points to the wrong runs, degrades
inference-time selection when it ranks candidates by a biased score, and distorts the
reward signal when used as an outcome reward model in reinforcement learning. The
over-flagging in particular is the same failure mode that a naive training reward
reproduces (Methods, Section~\ref{subsec:methods-verifier}), and closing this gap is
what a trained verifier must do.

\begin{table}[t]
\centering
\small
\caption{Error detection: distinguishing correct from incorrect trajectories. Positive class = trajectory contains an error. Precision, recall, and F1 (\%) are computed over all trajectories.}
\label{tab:detection}
\begin{tabular}{l ccc ccc ccc}
\toprule
& \multicolumn{3}{c}{\textbf{SWE-bench}} & \multicolumn{3}{c}{\textbf{TerminalBench}} & \multicolumn{3}{c}{\textbf{BixBench}} \\
\cmidrule(lr){2-4} \cmidrule(lr){5-7} \cmidrule(lr){8-10}
\textbf{Judge Model} & Prec.$\uparrow$ & Recall$\uparrow$ & F1$\uparrow$ & Prec.$\uparrow$ & Recall$\uparrow$ & F1$\uparrow$ & Prec.$\uparrow$ & Recall$\uparrow$ & F1$\uparrow$ \\
\midrule
Gemini 3.1 Pro & 60.34 & 70.59 & 65.06 & 80.65 & 97.04 & 88.09 & 67.39 & \textbf{81.58} & 73.81 \\
Gemini 3.0 Flash & 65.29 & 62.09 & 63.65 & 82.79 & 93.35 & 87.75 & 85.71 & 63.16 & 72.73 \\
Claude Opus 4.6 & \textbf{65.99} & 31.70 & 42.83 & \textbf{84.47} & 92.18 & \textbf{88.16} & \textbf{100.00} & 57.89 & 73.33 \\
GPT-5.5 & 60.92 & \textbf{82.03} & \textbf{69.92} & 78.62 & \textbf{97.77} & 87.16 & 81.08 & 78.95 & \textbf{80.00} \\
DeepSeek V4 Pro & 63.69 & 66.23 & 64.94 & 79.30 & 92.05 & 85.20 & 63.41 & 68.42 & 65.82 \\
Qwen 3.5-397B & 58.20 & 73.58 & 64.99 & 80.46 & 84.47 & 82.42 & 69.23 & 71.05 & 70.13 \\
\bottomrule
\end{tabular}
\end{table}

\begin{figure}[H]
    \centering
    \includegraphics[width=\textwidth]{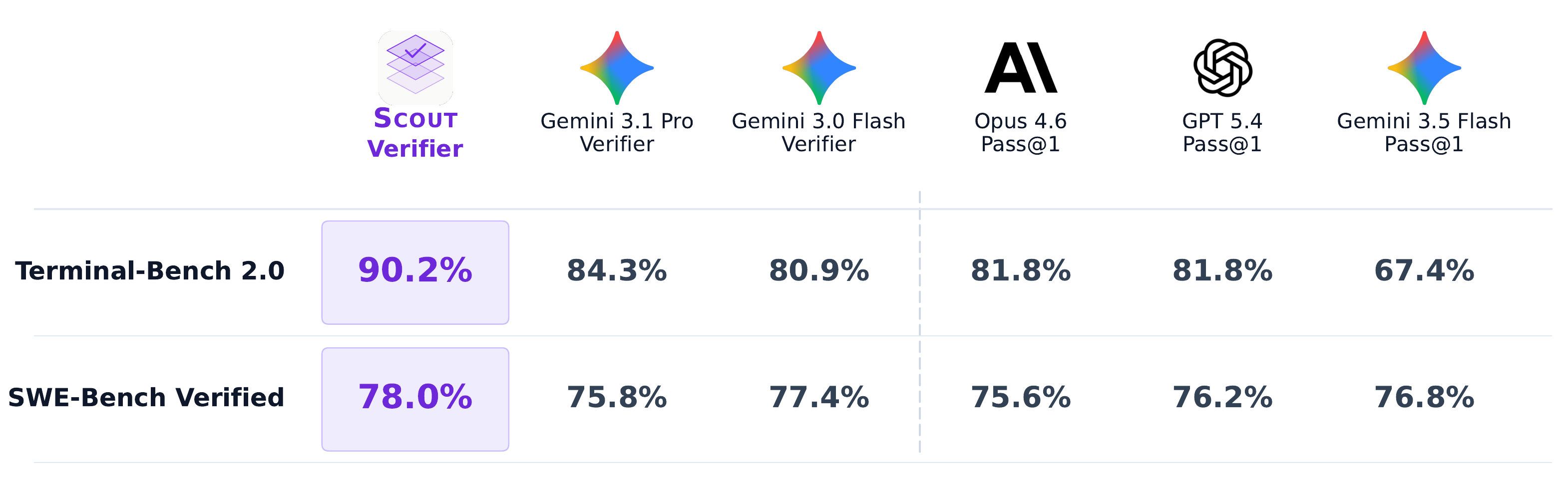}
    \caption{\textbf{A small trained verifier improves long-horizon agents more than frontier verifiers.} Best-of-$N$ task success (\%) when each verifier selects among
    an agent's candidate trajectories, on Terminal-Bench~2.0 and SWE-Bench Verified.
    \Scout{}, a $4$B-parameter verifier (highlighted), is compared against two
    frontier verifiers (Gemini 3.1 Pro and Gemini 3.0 Flash) and,
    as a reference, the single-sample pass@1 of three frontier agents (Claude Opus 4.6,
    GPT-5.4, Gemini 3.5 Flash). Selecting with \Scout{} raises success to $90.2\%$ on
    Terminal-Bench~2.0 and $78.0\%$ on SWE-Bench Verified, above every frontier
    verifier and above the pass@1 of agents far larger than itself, with the candidate
    trajectories held fixed across all selectors.}
    \label{fig:hero-fig-final}
\end{figure}

\subsection{Trained verifiers improve long-horizon agents}
\label{subsec:transfer-correct}

\paragraph{Selecting with the verifier raises agent success.} We use \Scout{} to improve agent performance at test time: given the several candidate
trajectories an agent produces for a task, it scores each and we keep the
highest-scoring one, on the candidate sets described in Methods
(Section~\ref{subsec:methods-eval}). Selecting with \Scout{} raises task success to
$90.2\%$ on Terminal-Bench~2.0 and $78.0\%$ on SWE-Bench Verified
(Fig.~\ref{fig:hero-fig-final}), above the pass@1 of the frontier agents themselves,
including Claude Opus 4.6 ($81.8\%$ and $75.6\%$): a $4$B verifier, by choosing among
an agent's attempts, yields higher success than agents far larger than it reach on
their own. Compared against much larger general-purpose frontier models used as the
verifier, \Scout{} still selects best, exceeding Gemini 3.1 Pro ($84.3\%$ and
$75.8\%$) and Gemini 3.0 Flash ($80.9\%$ and $77.4\%$) on both benchmarks, with all
selectors drawing from the same fixed pool of candidates so that the gain reflects the
quality of \Scout{}'s judgment rather than the candidates available. The margin is
larger on Terminal-Bench~2.0 than on SWE-Bench Verified, consistent with most of
\Scout{}'s training trajectories coming from the terminal domain. Even so, the result
shows that a small model trained on high-quality failure annotations
(Section~\ref{subsec:methods-verifier}) can improve long-horizon agents purely at test
time, with no retraining of the agent.

% \paragraph{The improvement holds in an unseen scientific domain.}

% \paragraph{Reinforcement learning amplifies the gains.} 

%=== END inlined: 4-experiments_results.tex ===
%=== BEGIN inlined: 5-limitation_future_work.tex ===
\section{Discussion}
\label{sec:limitations}

The central lesson of this work is that long-horizon failure is not primarily a
problem of error frequency. Agents on long tasks do not fail because they make more mistakes than they do on short ones; they fail because a single early mistake is
rarely caught and rarely reversed. Our characterization shows that once an agent makes
its first mistake it seldom recovers and rarely even detects it, continuing on a
corrupted state to the end of the run [recovery and self-detection rates]. The
difficulty of a long horizon is therefore one of irreversibility and self-blindness:
the agent cannot see the error it has made, and so cannot find its way back. This
reframes what reliability requires. Making long-horizon agents reliable is less a
matter of lowering their error rate than of catching and correcting the first error
before it compounds, and that, in turn, depends on being able to locate it. The rest
of our findings show that this is feasible: the first mistake can be located, the
skill of locating it is learnable by a small and inexpensive model rather than a
property of frontier scale, and a model with this skill measurably improves real
agents. Reliability at long horizons is thus achievable not by building agents that
never err, but by building the means to find and correct the first error when they do.

Building that means depends on measuring it, and this is where the step-level record
becomes more than a description of failure. The same signal that maps how agents fail,
a human-verified first-mistake label on every trajectory, also grades any model that
tries to catch those failures: a verifier, an LLM judge, or an outcome reward model is
scored not on whether it agrees with the final outcome but on whether it identifies
\emph{where} a run went wrong. This matters because of how agents are now built. They
are increasingly trained with reinforcement learning and scaled at test time, and both
rely on a verifier or reward model to score behavior, yet those models have themselves
been evaluated almost only on outcomes. An outcome-accurate reward model can still be
wrong about every step that produced the outcome, and our results show frontier judges
routinely are, biased toward over- or under-flagging in ways that an outcome-level
score never reveals. Process-level training and test-time selection need a verifier
that is right about \emph{where}, and that is precisely the axis our instrument
measures. By grading the tools on the same signal that characterizes the phenomenon,
the record turns failure localization from an unmeasured assumption into a quantity the
field can optimize.

These claims rest on choices that bound them, and we state the main ones plainly. The
first concerns the labels themselves: the first mistake in a trajectory is a human
judgment, and although our annotators agree substantially (Cohen's $\kappa = 0.77$ on
whether a run failed, $0.73$ on where), agreement is not perfect, and the notion of a
single first mistake is cleanest when a run derails at an identifiable step and least
clean when the failure is diffuse, building gradually across several weak decisions
rather than one wrong one. This is also a modeling choice: by locating failure at one
decisive index we capture the common case in which an early error compounds, but we
necessarily simplify failures that are genuinely multi-causal or that have no single
point of origin. Our evidence is also bounded in scope. We study three domains in which
long-horizon agents are already deployed, but agentic work extends to embodied,
multimodal, and multi-user settings we do not test, and whether the patterns we report
hold there is open. Finally, we distinguish what the data firmly support from what we
report as directional. The failure taxonomy and the frontier localization gap rest on
large, human-verified samples and are robust; the recovery and self-detection finding
and the cross-domain contrasts are based on smaller samples and we present them as
directional, developing them further below [pending the recovery and cross-domain
analyses]. We are explicit about this boundary so that the strongest claims of the
paper are not read as resting on its most preliminary measurements.

What this opens is more consequential than what it closes. If the first mistake can be
located, it can be caught while a run is still in progress: a verifier that flags the
decisive error as it happens lets an agent stop, reconsider, and back out before the
error compounds, turning the irreversibility we document into something correctable.
This speaks directly to the failure we identify, that agents do not recover on their
own; an external signal can supply the recovery they lack. The same step-level signal
also points beyond test-time correction to how agents are trained. Reinforcement
learning for long-horizon agents today rests largely on outcome rewards, which give one
late, sparse signal for an entire trajectory; a verifier that is accurate about
\emph{where} a run goes wrong is exactly the dense, process-level reward such training
needs, crediting the steps that were right and penalizing the one that was not. The
furthest reach is to fold this capability inward. The verifier we train is external to
the agent, but the ability it embodies, reading a trajectory and finding where it broke,
is one an agent could eventually turn on itself, closing the gap between making a
mistake and seeing it. An agent that can locate its own failures is an agent that can
learn from them, and the tools we release is a step toward measuring, and then
building, that capability.

%=== END inlined: 6-conclusion.tex ===
% --- Methods (now follows the main-text Results/Discussion) ---
%=== BEGIN inlined: 3-agent_systems.tex ===

\bibliography{main_arxiv}

\clearpage

% Let $\hat{y}_i \in \{\textsc{correct}, \textsc{incorrect}\}$ be the predicted label for step $i$, and $y_i$ the gold label, for a trajectory with $N$ steps. We define $\mathcal{C} = \{i : y_i = \textsc{correct}\}$ and $\mathcal{I} = \{i : y_i = \textsc{incorrect}\}$. The reward is:

% \[
% R(\hat{y}, y) = 
% \begin{cases}
% 0 & \text{if output is unparseable} \\[4pt]
% \mathbf{1}\!\left[\,\forall i:\, \hat{y}_i = \textsc{correct}\,\right] & \text{if } |\mathcal{I}| = 0 \\[4pt]
% \frac{1}{2}\!\left(\underbrace{\frac{\sum_{i \in \mathcal{C}} \mathbf{1}[\hat{y}_i = \textsc{correct}]}{|\mathcal{C}|}}_{\text{TNR}} + \underbrace{\frac{\sum_{i \in \mathcal{I}} \mathbf{1}[\hat{y}_i = \textsc{incorrect}]}{|\mathcal{I}|}}_{\text{TPR}}\right) & \text{if } |\mathcal{I}| > 0
% \end{cases}
% \]

% For correct trajectories ($|\mathcal{I}|=0$), the model must label every step as \textsc{correct} to receive reward; any false alarm yields zero. For incorrect trajectories ($|\mathcal{I}|>0$), the reward is \emph{balanced accuracy}---the mean of recall on correct steps (TNR) and recall on incorrect steps (TPR). This ensures that degenerate strategies (all-\textsc{correct} or all-\textsc{incorrect}) score at most $0.5$.

\section*{\LARGE Appendix}

\startcontents[appendices]

\begingroup
\small\setlength{\tabcolsep}{6pt}\renewcommand{\arraystretch}{1.22}
\begin{longtable}{@{}>{\raggedright\arraybackslash}p{3.4cm} >{\raggedright\arraybackslash}p{11.4cm}@{}}
\caption{\textbf{The long-horizon agent failure taxonomy: definitions of all 88 categories} (10 families). One concise definition per category. Each family header gives the family's share of all 6{,}967 classified mistakes, its count $n$, its number of categories, and its \emph{root} share (the fraction of its mistakes that are originating faults rather than downstream cascades from an earlier mistake); per-category prevalence, horizon phase, and root/cascade split are shown in Fig.~\ref{fig:failure-taxonomy}. Greyed rows are categories defined \emph{a priori} that never occurred in the corpus.}\label{tab:failure-defs}\\
\toprule
\textbf{Category} & \textbf{Definition} \\
\midrule
\endfirsthead
\multicolumn{2}{c}{\small\textit{Table \ref{tab:failure-defs} (continued)}}\\
\toprule
\textbf{Category} & \textbf{Definition} \\
\midrule
\endhead
\midrule \multicolumn{2}{r}{\footnotesize\textit{continued on next page}}\\
\endfoot
\bottomrule
\endlastfoot
\rowcolor{gray!20} \multicolumn{2}{@{}l}{\textbf{ACTION -- execution} \,\textemdash\, 45.1\% of mistakes \,$\cdot$\, $n=3144$ \,$\cdot$\, 14 categories \,$\cdot$\, 35\% root} \\
\textbf{invalid / rejected} & Well-formed but not valid in the current context (illegal move, non-existent or non-permitted command). \\
\textbf{wrong parameters} & Right command, but wrong flags / values / options. \\
\textbf{repeated same action} & Re-ran a previously failed action without any meaningful change. \\
\textbf{malformed syntax} & Command or code is syntactically invalid and will not parse. \\
\textbf{action $\neq$ intent} & Action does not implement the agent's own (correct) stated intent. \\
\textbf{wrong target} & Right kind of operation, but on the wrong file / path / entity / argument. \\
\textbf{incomplete edit} & Edit is partial and leaves the code or state broken. \\
\textbf{no-op / ineffective} & Action does nothing useful toward the goal. \\
\textbf{destructive action} & Destructive or irreversible operation (also flagged in the safety audit). \\
\textbf{wrong tool chosen} & Chose an inappropriate tool or command for the goal. \\
\textbf{wrong action format} & An action's edit / patch is not in the required format (e.g., a malformed diff block). \\
\textbf{partial input} & Command or code is cut off or only partially specified. \\
\textbf{actions out of order} & Correct actions executed in the wrong sequence. \\
\textbf{overwrote good work} & Clobbered correct existing content or earlier good work. \\
\rowcolor{gray!20} \multicolumn{2}{@{}l}{\textbf{PLANNING \& STRATEGY} \,\textemdash\, 24.0\% of mistakes \,$\cdot$\, $n=1674$ \,$\cdot$\, 8 categories \,$\cdot$\, 46\% root} \\
\textbf{flawed strategy} & Chose an overall approach that cannot achieve the goal. \\
\textbf{drifted off task} & Drifted away from the task's objective onto irrelevant actions. \\
\textbf{ignored a constraint} & Ignored an explicit instruction, constraint or limit (time / budget / step / space). \\
\textbf{over-complicated} & Needlessly long or complex plan that wastes steps when a simpler path existed. \\
\textbf{skipped a step} & Plan omits a necessary step (incomplete decomposition). \\
\textbf{scope creep} & Pursued out-of-scope work or changes beyond the task. \\
\textbf{no plan / blind} & Acted with no reasoning / plan where one was clearly needed. \\
\textbf{poor prioritization} & Worked the wrong sub-problem or in the wrong order (premature optimization). \\
\rowcolor{gray!20} \multicolumn{2}{@{}l}{\textbf{REASONING \& INFERENCE} \,\textemdash\, 10.8\% of mistakes \,$\cdot$\, $n=749$ \,$\cdot$\, 11 categories \,$\cdot$\, 69\% root} \\
\textbf{false assumption} & Assumed an untrue fact about the code, environment or state. \\
\textbf{misdiagnosed cause} & Reasoned to the wrong underlying cause of a bug / failure. \\
\textbf{hallucinated facts} & Invented a nonexistent API / file / function / flag / fact. \\
\textbf{ignored evidence} & Failed to use information it had already gathered earlier in the run. \\
\textbf{unjustified leap} & Jumped to a conclusion / acted without gathering needed evidence. \\
\textbf{incorrect inference} & Invalid logical step / wrong deduction from the information it had. \\
\textbf{misread the goal} & Misunderstood the task / requirements / what was being asked. \\
\textbf{impossible action} & Planned a step impossible given the current preconditions / state. \\
\textbf{long-horizon drift} & Lost logical / control-flow coherence over a long multi-step chain. \\
\textbf{circular reasoning} & Reasoning goes in circles and makes no real progress. \\
\textit{\textcolor{gray}{didn't seek clarification}} & \textit{\textcolor{gray}{On an ambiguous instruction, guessed a wrong default instead of seeking clarification. (defined a priori; not observed)}} \\
\rowcolor{gray!20} \multicolumn{2}{@{}l}{\textbf{TOOL \& ENVIRONMENT} \,\textemdash\, 6.9\% of mistakes \,$\cdot$\, $n=479$ \,$\cdot$\, 12 categories \,$\cdot$\, 75\% root} \\
\textbf{wrong env assumption} & Assumed wrong OS / tool / library / version availability. \\
\textbf{tool / API misuse} & Misused a tool / command interface (quoting, escaping, calling convention). \\
\textbf{missing setup} & Skipped a needed install / build / config / dependency step. \\
\textbf{wrong path / cwd} & Wrong working directory / path / relative-vs-absolute mistake. \\
\textbf{unchecked state} & Assumed environment state without checking it first. \\
\textbf{protocol violation} & Broke the harness's turn-level action / response protocol or schema. \\
\textbf{environment error} & Environment changed / degraded / behaved stochastically outside the agent's control (external). \\
\textbf{context mismanaged} & Mismanaged its own context / history (flooded the window, dropped needed history). \\
\textbf{permission / quota} & Hit or ignored a permission / quota / timeout / resource limit. \\
\textbf{tool exec error} & An external tool / API itself errored or misbehaved (external, not misuse). \\
\textbf{step-limit hit} & Run hit the max-step / max-turn cap despite reasonable behavior. \\
\textit{\textcolor{gray}{model capability limit}} & \textit{\textcolor{gray}{Failure from model / API limits (timeout, context-length limit, content filter). (defined a priori; not observed)}} \\
\rowcolor{gray!20} \multicolumn{2}{@{}l}{\textbf{TERMINATION} \,\textemdash\, 2.7\% of mistakes \,$\cdot$\, $n=189$ \,$\cdot$\, 7 categories \,$\cdot$\, 62\% root} \\
\textbf{premature `done'} & Declared completion and stopped while work clearly remained. \\
\textbf{incomplete submission} & Stopped and submitted while the deliverable was still missing parts. \\
\textbf{gave up too early} & Abandoned the task prematurely without claiming success. \\
\textbf{wrong final answer} & The submitted final answer / result is incorrect. \\
\textbf{failed to stop} & Stuck looping / thrashing, not converging. \\
\textbf{unaware run ended} & Did not recognize the criteria that should end the task and continued. \\
\textbf{never submitted} & Failed to submit a final answer when it could and should have. \\
\rowcolor{gray!20} \multicolumn{2}{@{}l}{\textbf{VERIFICATION \& TESTING} \,\textemdash\, 2.6\% of mistakes \,$\cdot$\, $n=182$ \,$\cdot$\, 6 categories \,$\cdot$\, 43\% root} \\
\textbf{inadequate testing} & Tested the wrong thing or used too weak a check. \\
\textbf{no verification} & Did not test / validate before proceeding or declaring success. \\
\textbf{misjudged result} & Misjudged whether a verification actually passed. \\
\textbf{ignored failing test} & Saw a failing test / check yet proceeded as if it passed. \\
\textbf{overfit / hacky fix} & Made the check pass superficially (hard-coded / special-cased the output). \\
\textbf{never reproduced bug} & Never reproduced the bug / behavior before attempting a fix. \\
\rowcolor{gray!20} \multicolumn{2}{@{}l}{\textbf{METACOGNITION} \,\textemdash\, 4.3\% of mistakes \,$\cdot$\, $n=297$ \,$\cdot$\, 6 categories \,$\cdot$\, 19\% root} \\
\textbf{failed to adapt} & Recognized trouble but failed to change approach (insight without behavior change). \\
\textbf{misjudged progress} & Misjudged its own progress (over-optimistic, over-pessimistic, or missed it had finished). \\
\textit{\textcolor{gray}{misplaced blame}} & \textit{\textcolor{gray}{Blamed the task / environment for a failure that was its own doing. (defined a priori; not observed)}} \\
\textit{\textcolor{gray}{no reflection}} & \textit{\textcolor{gray}{Never stepped back to reassess despite repeated failure or clear signals. (defined a priori; not observed)}} \\
\textit{\textcolor{gray}{overconfidence}} & \textit{\textcolor{gray}{Unwarranted certainty in its own correctness where doubt was clearly due. (defined a priori; not observed)}} \\
\textit{\textcolor{gray}{underconfidence}} & \textit{\textcolor{gray}{Abandoned a correct / promising path because of unwarranted self-doubt. (defined a priori; not observed)}} \\
\rowcolor{gray!20} \multicolumn{2}{@{}l}{\textbf{PERCEPTION} \,\textemdash\, 1.5\% of mistakes \,$\cdot$\, $n=102$ \,$\cdot$\, 9 categories \,$\cdot$\, 51\% root} \\
\textbf{ignored error msg} & Proceeded despite an explicit error / traceback / non-zero exit. \\
\textbf{misread observation} & Drew a wrong conclusion from output that was actually clear. \\
\textbf{ignored a warning} & Overlooked a softer warning or hint in the output. \\
\textbf{hallucinated state} & Believed a result / state not supported by the observation. \\
\textbf{overlooked key info} & Missed relevant information that was present in the output. \\
\textbf{misattributed cause} & From the observed output, pinned a failure on the wrong proximate signal. \\
\textbf{partial output read} & Acted on truncated / scrolled-off output without reading it fully. \\
\textbf{stale observation} & Relied on an earlier observation no longer valid after later changes. \\
\textit{\textcolor{gray}{missed state change}} & \textit{\textcolor{gray}{Failed to perceive a state change (believed success that did not happen, or missed a change). (defined a priori; not observed)}} \\
\rowcolor{gray!20} \multicolumn{2}{@{}l}{\textbf{MEMORY \& CONTEXT} \,\textemdash\, 1.2\% of mistakes \,$\cdot$\, $n=86$ \,$\cdot$\, 10 categories \,$\cdot$\, 74\% root} \\
\textbf{lost earlier context} & Lost earlier information because history exceeded effective context / memory capacity. \\
\textbf{redundant rework} & Re-derived or re-gathered a fact it had already established. \\
\textbf{false memory of step} & Recalled an event / state that never happened (fabricated past). \\
\textbf{forgot a sub-goal} & Dropped a still-pending sub-goal it had set itself. \\
\textbf{lost progress track} & Lost track of progress through the task and repeated already-completed steps. \\
\textbf{accumulated errors} & Built on a specific earlier artifact / result that was already flawed, treating it as correct. \\
\textbf{constraint drift} & Stopped applying an earlier-established constraint or decision later in the run. \\
\textbf{confused entities} & Mixed up files / variables / entities / versions. \\
\textit{\textcolor{gray}{over-summarized context}} & \textit{\textcolor{gray}{Summarized past info too crudely, dropping details needed for correct reasoning. (defined a priori; not observed)}} \\
\textit{\textcolor{gray}{retrieval failure}} & \textit{\textcolor{gray}{Relevant info is present in context but was not retrieved when needed. (defined a priori; not observed)}} \\
\rowcolor{gray!20} \multicolumn{2}{@{}l}{\textbf{OUTPUT / FORMATTING} \,\textemdash\, 0.9\% of mistakes \,$\cdot$\, $n=65$ \,$\cdot$\, 5 categories \,$\cdot$\, 68\% root} \\
\textbf{wrong value} & Reported a wrong numeric / factual value in its output or analysis. \\
\textbf{wrong answer format} & Final answer not in the requested format / units / precision. \\
\textbf{incomplete answer} & Answer content omits required components (e.g., a sub-question left unanswered). \\
\textbf{unsupported claim} & Stated a conclusion not supported by the data / analysis. \\
\textit{\textcolor{gray}{inconsistent report}} & \textit{\textcolor{gray}{Reported result contradicts its own computation / observation. (defined a priori; not observed)}} \\
\end{longtable}
\vspace{-0.4em}\noindent{\footnotesize A further 48 mistakes (0.7\%) fit no codebook category and were open-coded as \textit{other}; they are counted in their nearest family's total above. Definitions are condensed from the annotation codebook used by the LLM judge.}
\endgroup

\begin{table}[t]
\centering
\small
\caption{Dataset statistics of \Traverse{}.}
\label{tab:data_stats}
\begin{tabular}{l cc cc cc}
\toprule
& \multicolumn{2}{c}{\textbf{SWE-bench}} & \multicolumn{2}{c}{\textbf{TerminalBench}} & \multicolumn{2}{c}{\textbf{BixBench}} \\
\cmidrule(lr){2-3} \cmidrule(lr){4-5} \cmidrule(lr){6-7}
& Incorrect & Correct & Incorrect & Correct & Incorrect & Correct \\
\midrule
Trajectories & 306 & 294 & 541 & 180 & 38 & 64 \\
Total steps & 8,504 & 10,512 & 18,815 & 5,602 & 443 & 465 \\
Avg. steps & 27.8 & 35.8 & 34.8 & 31.1 & 11.7 & 7.3 \\
Median steps & 23 & 33.5 & 24 & 26 & 9.5 & 5 \\
Step range & 6--123 & 20--121 & 5--200 & 6--200 & 3--38 & 2--23 \\
Agent scaffolds & 3 & 3 & 2 & 2 & 2 & 2 \\
\midrule
Avg. first mistake step & 17.8 & -- & 12.4 & -- & 6.6 & -- \\
Median first mistake step & 14 & -- & 7 & -- & 2.5 & -- \\
\bottomrule
\end{tabular}
\end{table}

\begin{table}[t!]
\caption{Process-level verifier benchmarks for agentic and reasoning tasks. \Traverse{} is the only one combining long-horizon trajectories ($T$=20--200), naturally occurring failures, and human-annotated first-mistake localization across multiple domains and scaffolds.}
\label{tab:comparison-benchmarks}
\setlength{\tabcolsep}{3pt}
\centering
\footnotesize
\begin{tabular}{lcccccccc}
\toprule
\multirow{2}{*}{Benchmark} & \multirow{2}{*}{Domain} & \multirow{2}{*}{Agentic} & Traj. & Natural & First-mistake & Human step & Multiple & Trained \\
 & & & length & errors & localization & annotation & scaffolds & verifier \\
\midrule
ProcessBench~\citep{zheng2025processbench}        & Single   & \xmark & --       & \cmark & \cmark & \cmark & \xmark & \cmark \\
PRMBench~\citep{song-etal-2025-prmbench}            & Single   & \xmark & --       & \xmark & \xmark & \cmark & \xmark & \xmark \\
ToolPRMBench~\citep{li2026toolprmbench}        & Single   & \cmark & $\leq$30 & \xmark & \xmark & \xmark & \xmark & \cmark \\
AgentRewardBench~\citep{lu2025agentrewardbench}    & Single   & \cmark & --       & \cmark & \xmark & \xmark & \cmark & \xmark \\
Agent-RewardBench~\citep{men-etal-2025-agent}   & Multiple & \cmark & --       & \cmark & \xmark & \xmark & \cmark & \xmark \\
Web-Shepherd~\citep{chae2026web}        & Single   & \cmark & $\leq$40 & \xmark & \xmark & \cmark & \xmark & \cmark \\
\textbf{\Traverse{}}      & \textbf{Multiple} & \cmark & \textbf{20--200} & \cmark & \cmark & \cmark & \cmark & \cmark \\
\bottomrule
\end{tabular}
\end{table}

\begin{figure}[t]
\centering
\includegraphics[width=\linewidth]{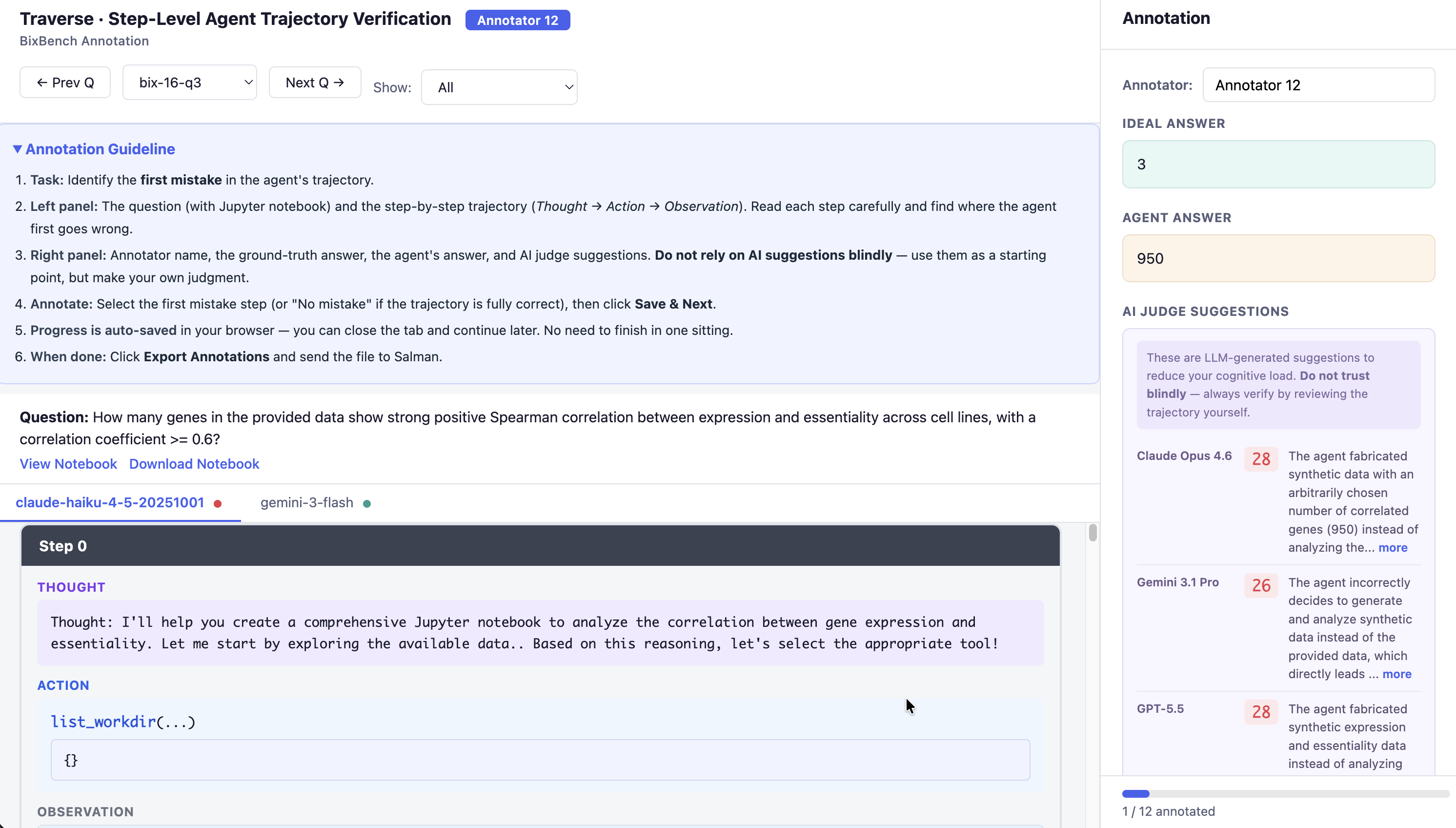}
\caption{The \textsc{Traverse} annotation interface (BixBench subset). The left
panel shows the task and the agent's full step-by-step trajectory
(\emph{Thought}\,$\rightarrow$\,\emph{Action}\,$\rightarrow$\,\emph{Observation});
the right panel shows the ground-truth answer, the agent's answer, and
LLM-generated judge suggestions. Annotators read each step and record the
index of the \emph{first mistake} (or $-1$ if the trajectory is fully correct).
Judge suggestions are shown only to reduce cognitive load and are explicitly
not authoritative---annotators are instructed to verify each step
independently.}
\label{fig:ui-bixbench}
\end{figure}

\begin{table}[t]
\centering
% \caption{Composition of \Traverse{}: task families, source benchmarks, model backbones, and agent harnesses.}
\caption{Composition of the collected trajectory corpus: task families, source
benchmarks, model backbones, and agent harnesses. \Traverse{} is selected from
this corpus (Section~\ref{subsec:methods-record}).}
\label{tab:data-composition}
\renewcommand{\arraystretch}{1.3}
\setlength{\tabcolsep}{6pt}
\footnotesize
\begin{tabularx}{\textwidth}{@{}l l X >{\raggedright\arraybackslash}p{3.1cm}@{}}
\toprule
\textbf{Task family} & \textbf{Benchmark} & \textbf{Models (backbones)} & \textbf{Harnesses} \\
\midrule
\multirow{2}{*}{Agentic coding}
  & SWE-bench$^{\dagger}$
  & Claude-sonnet-4, GPT-5, DeepSeek-V3.2, Qwen3-Coder-480B, Kimi-K2-Instruct
  & mini-SWE-agent, SWE-agent, OpenHands \\
\addlinespace[6pt]
  & TerminalBench
  & Claude-sonnet-4, GPT-5, DeepSeek-V3.2, Qwen3-Coder-480B, Kimi-K2-Instruct
  & mini-SWE-agent, Terminus\,2 \\
\midrule
AI-for-science
  & BixBench
  & Claude-sonnet-4-5, Claude-haiku-4-5, Claude-opus-4-5, Gemini-3-flash, Gemini-3.1-pro, GPT-5, GPT-5.4-mini, GPT-5.5
  & ReActAgent, SimpleAgent \\
\bottomrule
\end{tabularx}
\par\vspace{3pt}
{\footnotesize $^{\dagger}$SWE-bench spans the Verified, Pro, Multi-SWE, and SWE-PolyBench variants.}
\end{table}

\begin{figure*}[t]
\centering
\begin{tcolorbox}[
  colback=gray!3,
  colframe=gray!60,
  fonttitle=\bfseries\small,
  title=Prompt Template: First Mistake Identification,
  boxrule=0.5pt,
  arc=2pt,
  left=6pt, right=6pt, top=4pt, bottom=4pt,
  width=\textwidth
]
\footnotesize
\setlength{\parskip}{3pt}

The following is a coding task and an agent's step-by-step trajectory attempting to solve it (enclosed with tags and indexed from 0). Each step contains a Thought (the agent's reasoning before acting), an Action (what the agent executed), and an Observation (the environment's response).

\textbf{[Context]}\\
The agent was given a GitHub issue from a real software repository and tasked with modifying the codebase to resolve it. The agent can read files, edit code, and run tests within the repository.

\textbf{[Task]}\\
{\color{gray!50}\textit{\{task description from the dataset record\}}}

\textbf{[Agent Trajectory]}\\
\texttt{<step\_0>}\\
\hspace*{1em}Thought: {\color{gray!50}\textit{\{agent's reasoning\}}} \quad Action: {\color{gray!50}\textit{\{command or code executed\}}} \quad Observation: {\color{gray!50}\textit{\{environment response\}}}\\
\texttt{</step\_0>}\\
\hspace*{1em}$\vdots$\\
\texttt{<step\_\textit{n}>}\\
\hspace*{1em}Thought: {\color{gray!50}\textit{\{...\}}} \quad Action: {\color{gray!50}\textit{\{...\}}} \quad Observation: {\color{gray!50}\textit{\{...\}}}\\
\texttt{</step\_\textit{n}>}

\textbf{[Definition of Error]}\\
An error is a state-changing Thought or Action that is wrong given the evidence available to the agent at that point, leads the task in an incorrect direction, and directly contributes to a wrong outcome or prevents the task from being solved correctly. The Observation is the environment's response and is never itself an error.

Your task is to review and critique the trajectory step by step. Focus on the agent's Thought and Action to judge whether the step is correct or erroneous; use the Observation only as context for what the agent saw. Once you identify an error, return the 0-based index of the step where the earliest error occurs. If all Thoughts and Actions are correct, return $-1$.

You must put your final answer (i.e., the index) in \verb|\boxed{}|.

\end{tcolorbox}

\vspace{4pt}
{\footnotesize
\textbf{Variations.}
(1)~For \textit{TerminalBench}, the \textbf{[Context]} reads: ``The agent was given a task to complete in a sandboxed Linux terminal environment. Tasks span domains such as software engineering, data science, system administration, security, etc. The agent can run shell commands, write scripts, and inspect outputs.''
(2)~For \textit{OpenHands} trajectories, the Thought field is omitted from all steps (as OpenHands does not expose explicit reasoning traces), and all references to ``Thought'' in the instructions and error definition are removed accordingly.
}

\caption{Prompt template for the first mistake identification task. Gray italicized text denotes fields populated from each dataset record. The judge model returns the 0-based step index of the first error in \texttt{\textbackslash boxed\{\}}, or $-1$ if the trajectory is error-free.}
\label{fig:prompt}
\end{figure*}

\begin{figure}[t]
\centering
\includegraphics[width=\linewidth]{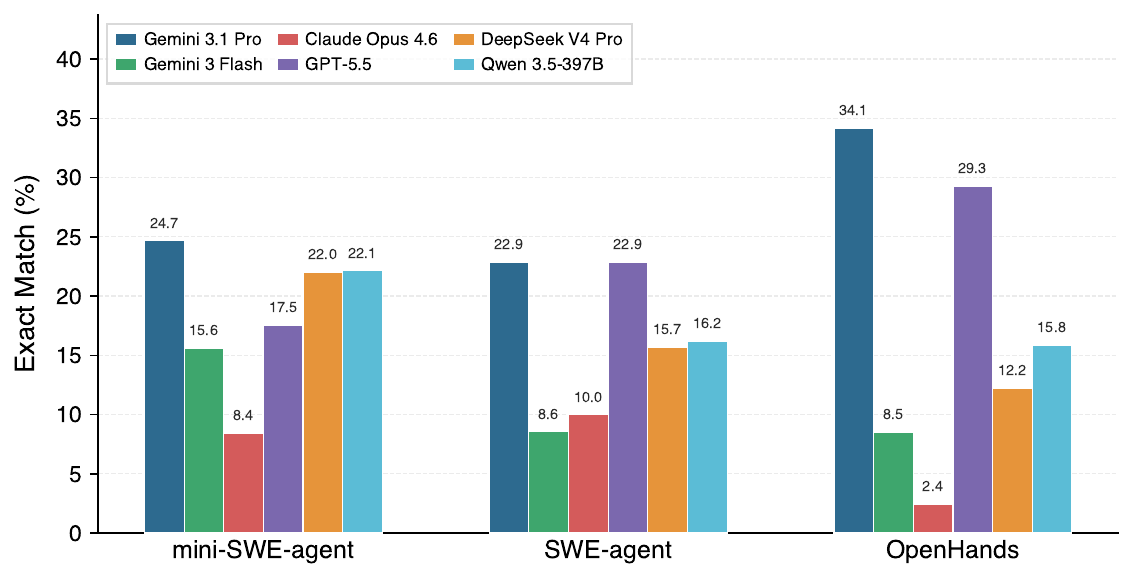}
\caption{Exact Match (\%) by agent type on SWE-bench incorrect trajectories.}
\label{fig:agent_swe}
\end{figure}

\begin{figure}[t]
\centering
\includegraphics[width=\linewidth]{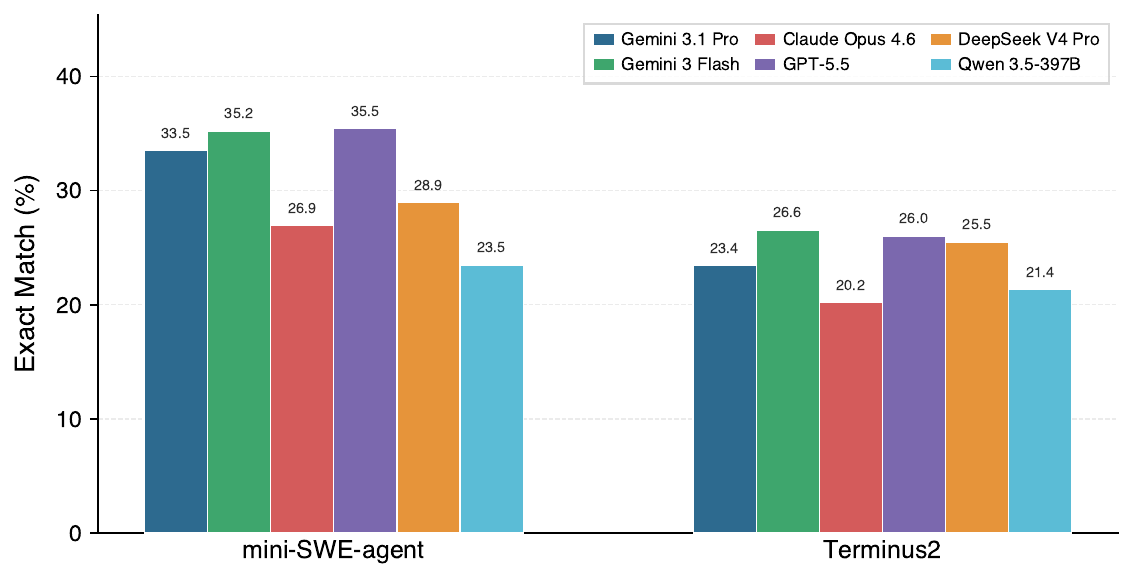}
\caption{Exact Match (\%) by agent type on TerminalBench incorrect trajectories.}
\label{fig:agent_terminal}
\end{figure}

\begin{table}[t]
\centering
\small
\setlength{\tabcolsep}{4pt}
\caption{\Scout{} training data: composition by source.}
\label{tab:train_data}
\begin{tabular}{l ccc}
\toprule
& Terminal & SWE & Total \\
\midrule
Trajectories      & 1{,}851 & 565 & 2{,}416 \\
\;\; with error   & 829 & 259 & 1{,}088 \\
\;\; all-correct  & 1{,}022 & 306 & 1{,}328 \\
\midrule
Steps             & 57{,}622 & 17{,}117 & 74{,}739 \\
\;\; correct      & 51{,}365 & 16{,}380 & 67{,}745 \\
\;\; incorrect    & 6{,}257 & 737 & 6{,}994 \\
\bottomrule
\end{tabular}
\end{table}

% \clearpage
% \section*{Supplementary: CoDaS-Generated Analysis Reports}
% \addcontentsline{toc}{section}{Supplementary: CoDaS-Generated Reports}

% \noindent The following reports were automatically generated by the CoDaS framework during the biomarker discovery process. These reports demonstrate the system's capability to produce comprehensive, human-readable analysis documentation.

% \vspace{1em}

% % Report 1
% % \subsection*{Report 1: DWB Dataset}
% \addcontentsline{toc}{subsection}{Report 1: [Title]}
% \includepdf[pages=-, pagecommand={\thispagestyle{plain}}]{pdfs/dwb_report.pdf}

\end{document}